\documentclass[10pt,letterpaper,logo,onecolumn,teaser]{planstyle}
\usepackage[numbers,sort&compress]{natbib}
\usepackage{makecell,threeparttable,multirow,wrapfig,subcaption,marvosym}
\hypersetup{hidelinks}
\fancypagestyle{firststyle}{%
  \fancyhf{}%
  \fancyfoot[L]{\footerfont\the\correspondingauthor}%
}

\usepackage{adjustbox}

\title{\raisebox{-0.6ex}{\includegraphics[height=3.8ex]{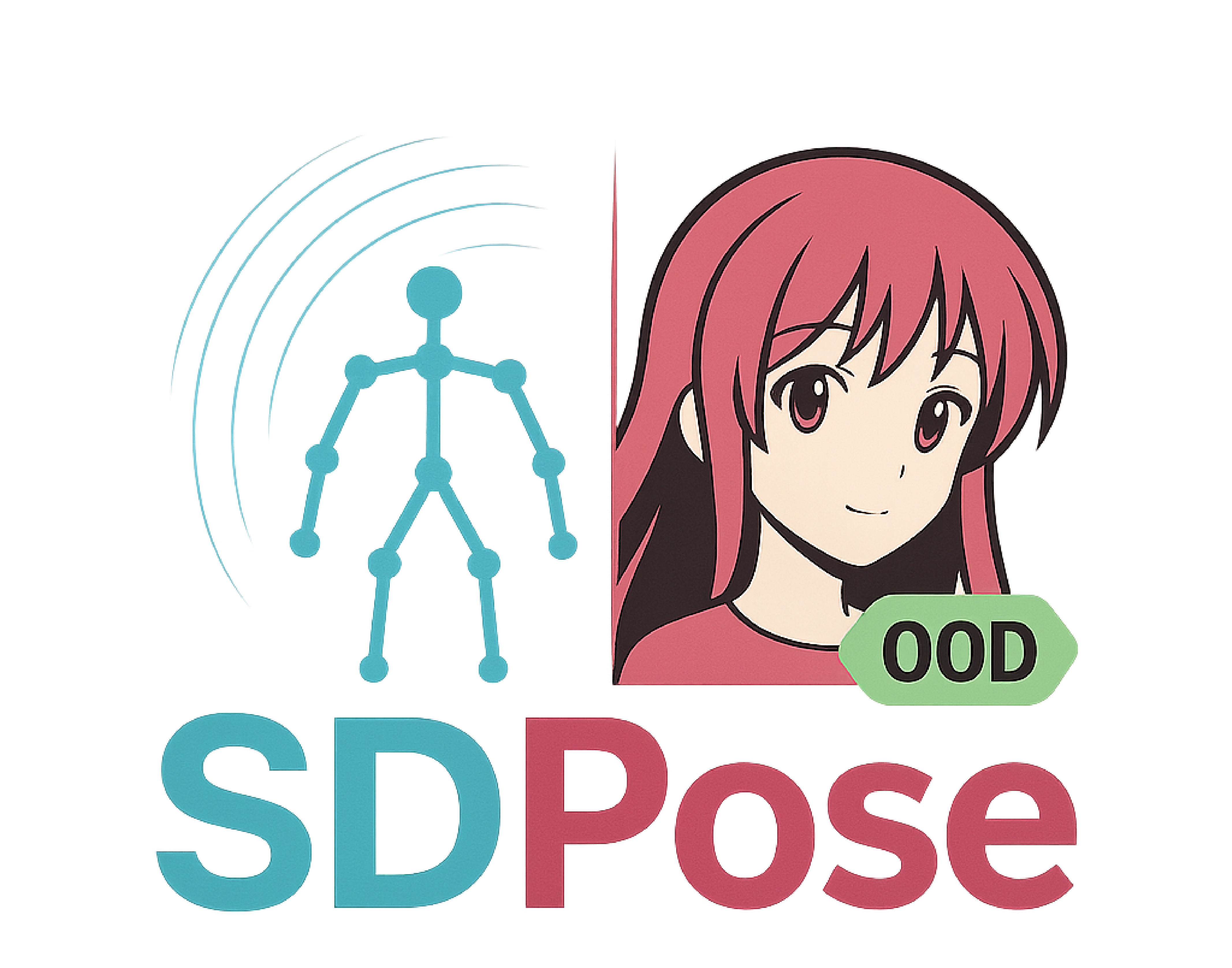}}\hspace{0.2em}SDPose: Exploiting Diffusion Priors for Out-of-Domain and Robust Pose Estimation}
\author{\begin{tabular}{c}
\textbf{Shuang Liang$^{1,5\dagger}$, Jing He$^{3\dagger}$, Lejun Liao$^{2}$, Chuanmeizhi Wang$^{1}$,}\\
\textbf{Guo Zhang$^{1,4}$, Yingcong Chen$^{3,6}$, Yuan Yuan$^{2*}$}
\end{tabular}}
\affil{
$^{1}$Rama Alpaca Technology Company\quad $^{2}$Boston College\quad $^{3}$HKUST (GZ)\\
$^{4}$Research Institute of Tsinghua University in Shenzhen\quad $^{5}$HKU\quad $^{6}$HKUST
}
\correspondingauthor{\begin{tabular}{@{}l@{}}
\small $^{*}$ Corresponding author.\\
\small $^{\dagger}$ Equal contribution. Shuang Liang's work was done during an internship at Rama Alpaca Technology.
\end{tabular}}

\renewcommand{\insertteaserfigure}{
\begin{center}
\captionsetup{type=figure}
  \centering
  \includegraphics[width=0.9\linewidth]{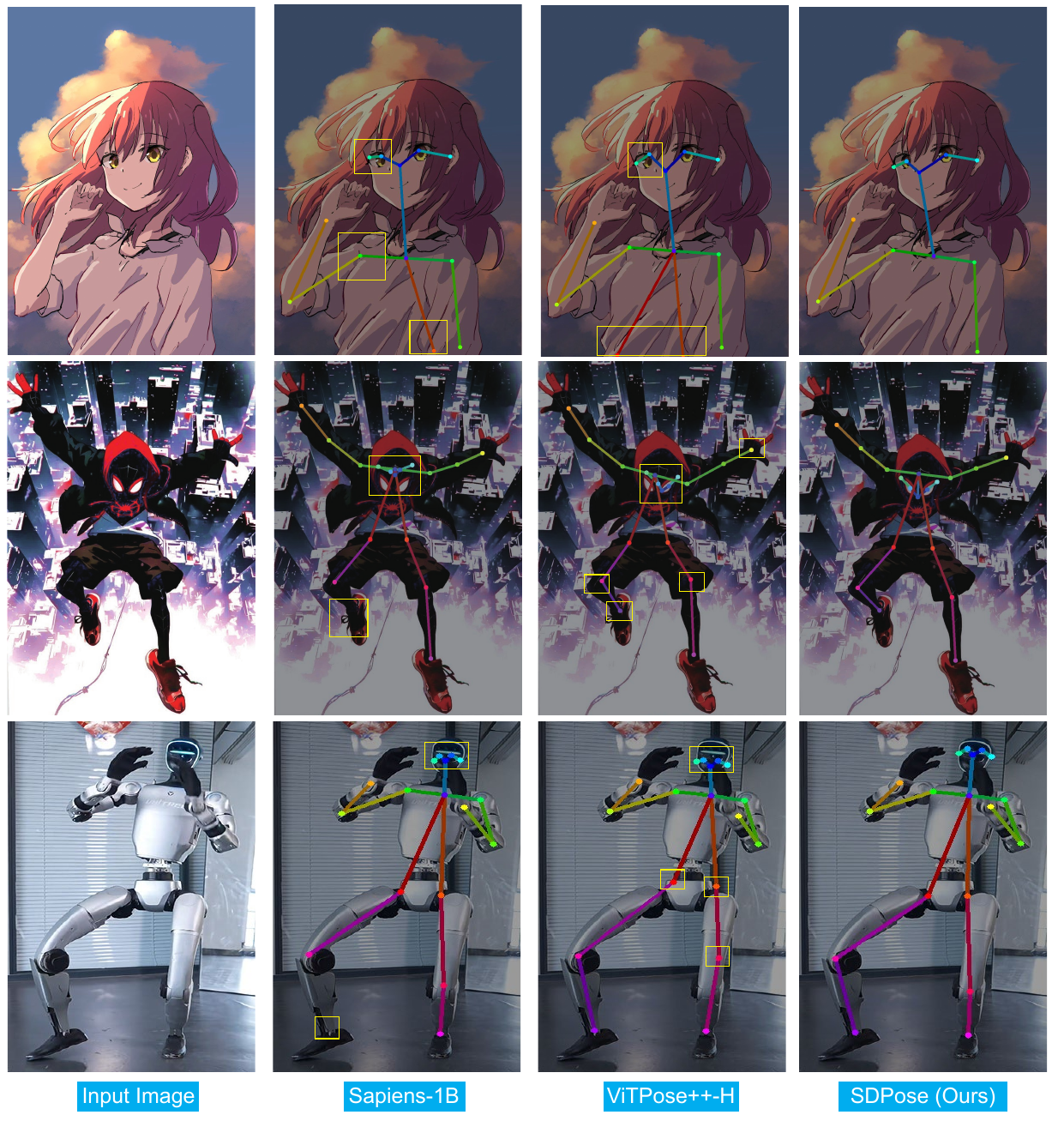} % 留一点边距防止溢出
  \caption{\textbf{SDPose for OOD pose estimation.} On stylized paintings and images of humanoid robots, SDPose produces more accurate predictions than Sapiens and ViTPose++-H. Baseline failure cases are highlighted in (\emph{yellow boxes}).}
  \label{fig:teaser}
\end{center}
}
\begin{document}
\begin{abstract}

Pre-trained diffusion models provide rich latent features across U-Net levels and are emerging as powerful vision backbones. While prior works such as Marigold and Lotus repurpose diffusion priors for dense geometric perception tasks such as depth and surface normal estimation, their potential for cross-domain human pose estimation remains largely unexplored. Through a systematic analysis of latent features from different upsampling levels of the Stable Diffusion U-Net, we identify the levels that deliver the strongest robustness and cross-domain generalization for pose estimation. Building on these findings, we propose \textbf{SDPose}, which (i) extracts U-Net features from the selected upsampling blocks, (ii) fuses them with a lightweight feature aggregation module to form a robust representation, and (iii) jointly optimizes keypoint heatmap supervision with an auxiliary latent reconstruction loss to regularize training and preserve the pre-trained generative prior. To evaluate cross-domain generalization and robustness, we construct COCO-OOD, a COCO-based benchmark with four subsets: three style-transferred splits to assess domain shift and one corruption split (noise, weather, digital artifacts, and blur) to test robustness. With data-efficient fine-tuning, SDPose achieves performance comparable to Sapiens on COCO, and establishes the best zero-shot performance on cross-domain datasets. Code is available at \url{https://github.com/T-S-Liang/SDPose-OOD}.

\end{abstract}
\maketitle

\section{Introduction}
\label{sec:intro}

\noindent With the recent rise of embodied AI, video generation \cite{kim2024tcan, hu2024animate}, and 3D asset rendering, the need for cross-domain and robust human pose estimation has become critical in robotics as well as in film, animation, and game production. Although recent advances on academic benchmarks such as MS COCO~\cite{lin2014microsoft} using models such as DWPose~\cite{yang2023effective}, RTMPose~\cite{jiang2023rtmpose} and OpenPose~\cite{martinez2019openpose}, as well as approaches leveraging large pre-trained backbones such as ViTPose~\cite{xu2022vitpose, xu2023vitpose++} and Sapiens~\cite{khirodkar2024sapiens}, have achieved strong in-domain accuracy, they often exhibit severe performance degradation under domain shifts and require substantial fine-tuning efforts.

Recently, pre-trained diffusion models such as Stable Diffusion~\cite{rombach2022high} have emerged as strong vision backbones.
While a growing body of work has shown that diffusion priors can be repurposed for 3D generation~\cite{cheng2023sdfusion,lin2025kiss3dgen,long2024wonder3d}, segmentation~\cite{karmann2025repurposing}, and dense geometric perception~\cite{ke2024repurposing,he2024lotus,he2025lotus}, their potential for robust, cross-domain human pose estimation remains underexplored.

\begin{figure*}[!t]
    \centering
    \includegraphics[width=\linewidth]{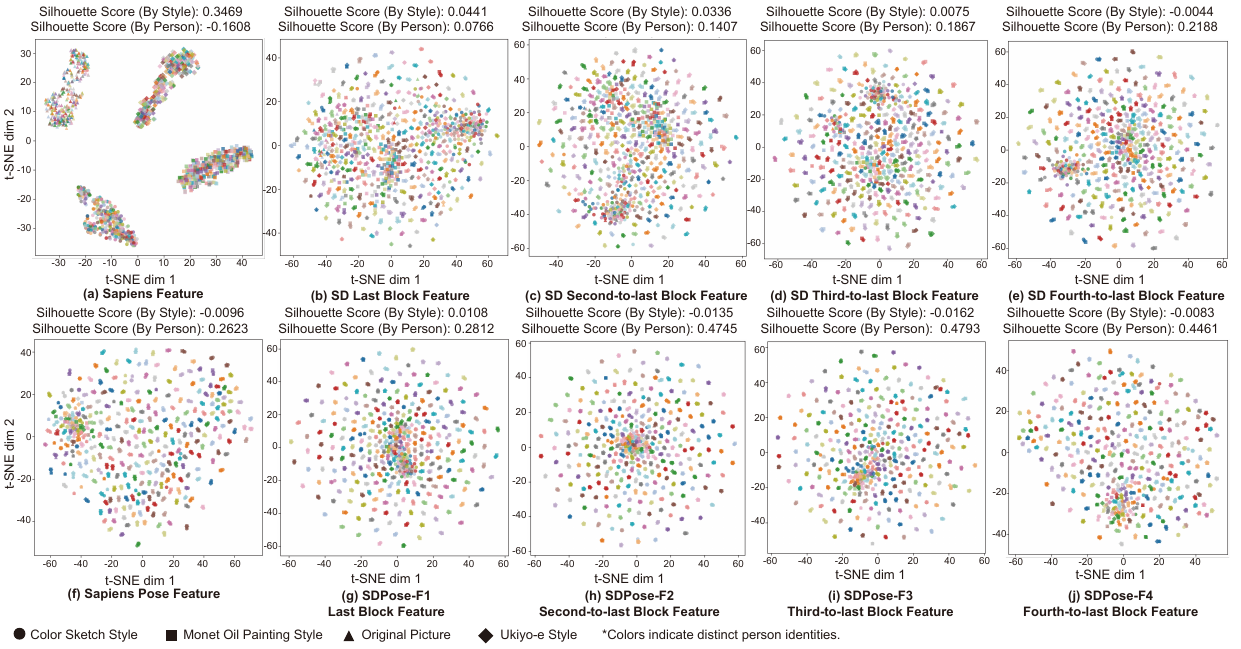}
    \caption{\textbf{t-SNE visualization of features from Sapiens ViT, Stable Diffusion U-Net blocks, Sapiens Pose, and SDPose across four visual domains.} Each point corresponds to an image sample; colors represent person instances and marker shapes denote artistic styles.}
    \label{fig:latent_space}
\end{figure*}

It is worth noting that we observe evidence from representation analysis that diffusion priors exhibit a different embedding-space structure from human-centric backbones under style shifts.
Using t-SNE~\cite{maaten2008visualizing} and silhouette scores~\cite{rousseeuw1987silhouettes} on 1200 person crops (300 COCO instances with three style-transferred variants), we compare features from pre-trained Sapiens-1B and the SD-v2 U-Net.
As shown in Fig.~\ref{fig:latent_space} (a--e), Sapiens features tend to form style-driven clusters, whereas SD U-Net features exhibit weaker style separation and stronger instance coherence across styles.
Across SD U-Net upsampling stages, the degree of cross-style consistency is not uniform: lower-resolution features typically encode more semantic cues, while higher-resolution features preserve more fine-grained details.
This motivates our core question: which SD U-Net upsampling levels are most reliable under domain shifts for pose estimation, and how can we exploit them under standard supervised training?

Concurrent efforts such as GenLoc~\cite{wang2025generalizable} and Diff-Tracker~\cite{zhang2024diff} repurpose generative priors for object keypoint localization across \emph{keypoint schemas} by learning condition embeddings and extracting cross-attention correlation maps. While effective for schema-flexible queries, this formulation is constrained by the spatial granularity of attention maps extracted at the diffusion latent resolution.

Targeting robust and cross-domain human pose estimation, we bypass the spatial bottlenecks of condition-driven attention.
Instead, we answer the aforementioned question by identifying the most reliable U-Net feature levels and decoding pose directly from their fused features under standard supervised training.
Building on this perspective, we present \textbf{SDPose} and a dedicated OOD benchmark to study the cross-domain generalization and robustness in pose estimation. Our contributions are as follows:

\textbf{(i) Systematic analysis of SD U-Net features across upsampling levels under domain shifts.}
We systematically analyze SD U-Net features from different upsampling levels via ablations and representation analysis (t-SNE and silhouette scores), identifying the most reliable levels for human pose estimation under cross-domain style shifts.

\textbf{(ii) Feature aggregation and prior-preserving fine-tuning.}
Guided by these findings, we select the appropriate feature levels, and introduce a feature aggregation module to fuse multi-level SD U-Net features, followed by a lightweight convolutional head for keypoint heatmap prediction. To preserve the generative priors of the pre-trained backbone during fine-tuning, we incorporate an auxiliary RGB reconstruction branch, which helps maintain domain-transferable features and improves cross-domain generalization.

\textbf{(iii) COCO-OOD benchmark.} To evaluate cross-domain generalization and robustness, we construct \textbf{COCO-OOD}, a COCO-based benchmark with four subsets: three style-transferred splits to assess domain shift, and one corruption split (noise, weather, digital artifacts, and blur) to test robustness.

On COCO~\cite{lin2014microsoft} and COCO-WholeBody~\cite{jin2020whole}, SDPose achieves in-domain performance comparable to Sapiens-1B/2B~\cite{khirodkar2024sapiens}. 
Under domain shifts (HumanArt~\cite{ju2023human} and COCO-OOD style-transfer subsets such as Monet and Ukiyo-e) as well as common corruptions, SDPose establishes strong zero-shot performance, surpassing Sapiens-2B while requiring shorter fine-tuning epochs and using a smaller parameter scale.
Beyond quantitative evaluation, we demonstrate that SDPose can serve as a zero-shot pose annotator for downstream controllable generation tasks, including ControlNet-based image synthesis and video generation, where it provides reliable and stable pose guidance.

\section{Related Works}

\subsection{Latent Diffusion Models}
Latent diffusion models (LDMs), built on DDPM and further advanced by ODE and SDE samplers~\cite{ho2020denoising,song2020score,song2020denoising,lu2022dpm,rombach2022high}, have gained traction over the past few years. Classic architectures, such as the UNet-based Stable Diffusion and Diffusion Transformers (DiT)~\cite{peebles2023scalable, esser2024scaling}, have demonstrated strong performance across diverse conditional generation tasks~\cite{zhang2023adding}. Pre-trained on large-scale datasets such as LAION-5B~\cite{schuhmann2022laion}, generative models like Stable Diffusion provide rich visual priors that can be effectively leveraged for a wide range of tasks. Recent advances in flow-matching~\cite{lipman2022flow, esser2024scaling, xie2024sana} further show that latent diffusion models can achieve high-quality synthesis with only a few sampling steps. These developments highlight the power of latent generative priors as a strong visual foundation.

\subsection{Leveraging Diffusion Priors for Prediction Tasks}

A growing body of work has explored repurposing pre-trained latent diffusion models as backbones for dense prediction.
Marigold~\cite{ke2024repurposing} adapts Stable Diffusion for monocular depth by fine-tuning the denoising U-Net with synthetic supervision.
Subsequent methods such as Lotus~\cite{he2024lotus, he2025lotus} and GenPercept~\cite{xu2024diffusion} further improve efficiency by replacing multi-step sampling with a deterministic one-step predictor for task outputs.
However, leveraging diffusion priors for robust, cross-domain human pose estimation remains comparatively underexplored.

Concurrent efforts such as GenLoc~\cite{wang2025generalizable} and Diff-Tracker~\cite{zhang2024diff} repurpose diffusion priors for object keypoint localization across \emph{keypoint schemas} by learning condition embeddings and extracting cross-attention correlation maps. While effective for schema-flexible queries, this formulation is constrained by the spatial granularity of attention maps extracted at the diffusion latent resolution, making it less suited for high-precision pose estimation under standard supervised training. In contrast, we systematically analyze features from different SD U-Net \emph{upsampling levels} and directly decode pose from fused features, bypassing condition-driven attention maps for robust and cross-domain pose estimation.

\begin{figure}[!t]
    \centering
    \includegraphics[width=0.65\linewidth,height=0.30\textheight,keepaspectratio]{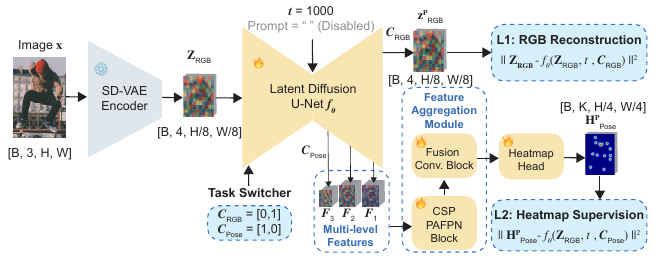}
        \caption{
        \textbf{Training pipeline of SDPose.}
        The input image $x$ is encoded by a pre-trained SD-VAE into $Z_{\mathrm{RGB}}$.
        A shared latent diffusion U-Net $f_{\theta}$ is conditioned by a task/class embedding $C$ (text prompt disabled).
        With $C_{\mathrm{RGB}}$, the model reconstructs $Z_{\mathrm{RGB}}$ via an $\ell_{2}$ latent reconstruction loss.
        With $C_{\mathrm{Pose}}$, we extract selected multi-level U-Net features, perform feature aggregation, and predict keypoint heatmaps supervised by a heatmap MSE loss.
        }
    \label{training_pipeline}
\end{figure}

\subsection{Human Pose Estimation}

Human pose estimation is a fundamental task in computer vision, yet remains challenging under severe domain shifts.
Early approaches predominantly relied on CNN backbones such as HRNet~\cite{sun2020bottom} and CSPNeXt~\cite{chen2024cspnext}, coupled with task-specific decoding heads.
Models such as RTMPose~\cite{jiang2023rtmpose} and DWPose~\cite{yang2023effective} achieve strong in-domain performance on benchmarks such as COCO and COCO WholeBody.
However, these models often generalize poorly under domain shifts from natural images to stylized domains such as artistic illustrations and anime.
More recently, methods built on large pre-trained backbones, such as ViTPose~\cite{xu2022vitpose, xu2023vitpose++} and Sapiens~\cite{khirodkar2024sapiens}, achieve state-of-the-art results on standard benchmarks, highlighting the benefit of foundation-scale pretraining.
Nevertheless, such approaches often incur high adaptation costs, requiring large task-specific data and lengthy training schedules to maintain performance under domain shift.
In this paper, we show that Stable Diffusion provides latent features that transfer well across styles, and that they can be adapted for human pose estimation via feature-level aggregation and prior-preserving fine-tuning, narrowing the cross-domain generalization gap with data-efficient fine-tuning.

\section{Preliminaries}

\subsection{Heatmap Representation and Unbiased Data Processing (UDP)}

Let $(x_i,y_i)$ denote the $i$-th ground-truth keypoint in an $H\times W$ image. 
The standard heatmap representation encodes each keypoint as
\begin{equation}
H_i(u,v)=\exp\!\left(-\frac{(u-x_i)^2+(v-y_i)^2}{2\sigma^2}\right),
\end{equation}
\begin{equation}
(\hat x_i,\hat y_i)=\arg\max_{u,v} H_i(u,v).
\end{equation}
While widely adopted, this discrete pixel-space formulation suffers from quantization bias: 
predicted coordinates become misaligned under flips, scales, or rotations since the argmax operation only yields integer positions. 
To address this issue, we adopt the Unbiased Data Processing (UDP) method~\cite{huang2020devil}, which removes quantization bias by estimating keypoints in a continuous domain. 
Following common practice, the heatmap is generated at one-quarter resolution of the input image, which balances localization accuracy with computational efficiency.
\subsection{Parameterization for Latent Diffusion Model}

Traditional latent diffusion models (LDMs)~\cite{ho2020denoising} adopt the $\epsilon$-prediction parameterization, 
where the denoiser $f_{\theta}$ is trained to predict the Gaussian noise $\epsilon_t$ added at timestep $t$: 
\begin{equation}
\hat{\epsilon}_t = f_\theta(z_t, t),
\end{equation}
with $z_t$ denoting the noisy latent at step $t$. The clean latent $x_0$ can then be recovered by
\begin{equation}
\hat{x}_0 = \frac{z_t - \sqrt{1 - \alpha_t}\,\hat{\epsilon}_t}{\sqrt{\alpha_t}},
\end{equation}
where $\alpha_t = \prod_{s=1}^t (1-\beta_s)$ is the cumulative product of the noise schedule.

However, Lotus \cite{he2024lotus} shows that \(\epsilon\)-prediction injects unnecessary stochastic variation, which accumulates across multiple denoising steps, degrading dense prediction quality. Lotus therefore advocates a \textbf{deterministic} adaptation, directly predicting the clean annotation latent \(x_0\) in a single step:
\begin{equation}
\hat{x}_0 = g_\theta(z_T),
\end{equation}
where \(T\) is a fixed timestep and \(g_\theta\) is fine-tuned U-Net applied once. This formulation eliminates the prediction variance introduced by multiple-step denoising, simplifies optimization, and significantly accelerates inference. In our approach, we similarly avoid the diffusion chain and adopt this \(x_0\)-prediction design.

\begin{figure}[!t]
    \centering
    \includegraphics[width=0.65\linewidth,height=0.30\textheight,keepaspectratio]{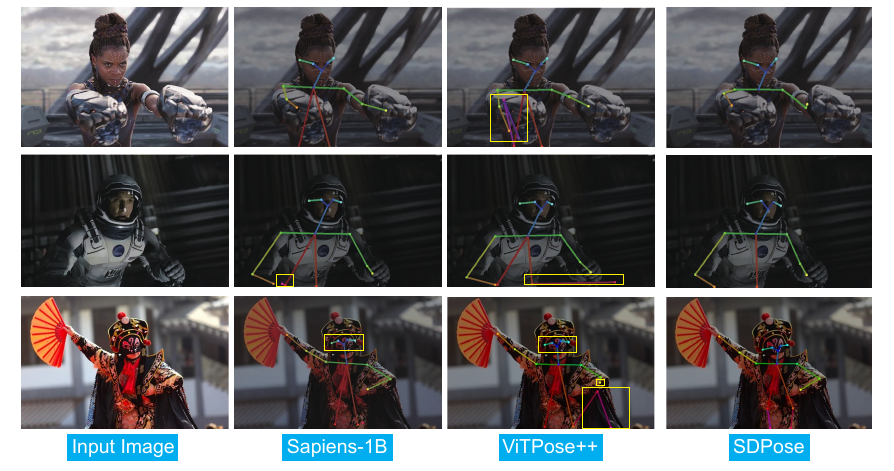}
    \caption{\textbf{Qualitative results on real-world photographs.} The yellow boxes highlight regions where baselines fail to predict accurate poses.}
    \label{fig:real_world}
\end{figure}

\section{Methodology}
\label{sec:method}

\subsection{Leveraging Multi-level SD Latent Features for Pose Estimation}
Given an input image, we encode it into the latent space using a frozen SD-VAE encoder and feed the latent to the SD U-Net.
From the U-Net upsampling hierarchy, we denote the candidate features from different upsampling blocks as $\{F_1,F_2,F_3,F_4\}$ (from shallower to deeper upsampling blocks).
Rather than using all feature levels, we perform a level-wise study (Sec.~\ref{sec:single-level-prob}) to identify the features that are both discriminative for keypoint prediction and stable under style shifts.
Based on this study, we select a robust subset $\mathcal{F}\subset\{F_1,F_2,F_3,F_4\}$ for subsequent feature aggregation and heatmap prediction.

\subsection{Feature Aggregation}

Relying on a single feature level can be suboptimal: deeper features tend to be more stable under style shifts, whereas shallower features retain finer spatial details for localization.
To exploit complementary cues while preserving generalization under domain shift, we introduce an aggregation module that fuses a selected subset of features $\mathcal{F}\subset\{F_1,F_2,F_3,F_4\}$ along the SD U-Net upsampling blocks (the selection rule is determined by the level-wise study in Sec.~\ref{sec:single-level-prob}).
In practice, we implement the aggregation with a lightweight PAFPN-style~\cite{lyu2022rtmdet} design that performs top-down and bottom-up fusion across feature levels.
The fused feature is then forwarded to a lightweight convolutional head~\cite{xiao2018simple} to predict keypoint heatmaps.
Ablations comparing single-level features and aggregated representations are provided in Sec.~\ref{sec:fusion-design}.

\subsection{Prior-preserving Auxiliary Reconstruction}
\label{sec:aux-rgb}

To preserve the generative priors and mitigate overfitting to the pose domain, we adopt the \textit{Detail Preserver} strategy~\cite{he2024lotus} by introducing an auxiliary RGB latent reconstruction objective (Fig.~\ref{training_pipeline}).
Concretely, we use a task/class embedding $C\in\{C_{\text{RGB}},C_{\text{Pose}}\}$ to switch the behavior of the shared denoising U-Net $f_{\theta}$.
When $C_{\text{RGB}}$ is provided, the network predicts the reconstructed latent $Z^{P}_{\text{RGB}}=f_{\theta}(Z_{\text{RGB}},t,C_{\text{RGB}})$ and is supervised by an $\ell_2$ reconstruction loss.
When $C_{\text{Pose}}$ is provided, we extract multi-scale U-Net features, fuse them, and predict keypoint heatmaps $H^{P}_{\text{Pose}}$, supervised by a heatmap MSE loss.
The overall objective is
\begin{equation}
\mathcal{L}
= \underbrace{\left\| Z_{\text{RGB}} - f_{\theta}(Z_{\text{RGB}}, t, C_{\text{RGB}}) \right\|_2^2}_{\mathcal{L}_{\text{RGB}}}
+ \underbrace{\left\| H_{\text{Pose}} - H^{P}_{\text{Pose}} \right\|_2^2}_{\mathcal{L}_{\text{Pose}}}.
\end{equation}
We adopt a single-step deterministic setting and fix the timestep to $t{=}1000$ for the U-Net forward used in both branches.
\subsection{Inference}

During inference, the input RGB image \(x\) is encoded by the SD-VAE into the latent representation $Z_{\text{RGB}}$. The latent diffusion U-Net then performs a single-step regression with the timestep fixed at \(t = 1000\), using the class label \(C_{\text{Pose}}\) to execute the pose estimation task. The text condition is disabled by feeding an empty text embedding to the U-Net.

\begin{table*}[t]
\caption{\textbf{Quantitative comparison across COCO, HumanArt, COCO-OOD Monet, COCO-OOD Ukiyo-e, and COCO-OOD Corruption.}
All models are trained on COCO. HumanArt~\cite{ju2023human} only provides 17-keypoint annotations (no WholeBody labels are available).
GenLoc~\cite{wang2025generalizable} results on WholeBody and certain COCO-OOD subsets are unavailable because the model and code have not been publicly released.
For clarity, we only show comparisons with Sapiens and GenLoc in the main paper: Sapiens represents the current state-of-the-art for 2D pose estimation, while GenLoc leverages diffusion generative priors for keypoint localization.
Full quantitative comparisons across various models are provided in the supplementary.}
\label{tab:quantitative}
\centering
\renewcommand{\arraystretch}{1.2}
\setlength{\tabcolsep}{5pt}
\footnotesize

\begin{adjustbox}{max width=\linewidth,center}

\begin{tabular}{
    c!{\vrule width .6pt}      % Model Variant
    c!{\vrule width .6pt}      % Model
    c!{\vrule width .6pt}      % Backbone
    c!{\vrule width .6pt}      % Params
    c c!{\vrule width .6pt}    % COCO
    c c!{\vrule width .6pt}    % HumanArt
    c c!{\vrule width .6pt}    % COCO-OOD Monet
    c c!{\vrule width .6pt}    % COCO-OOD Ukiyo-e
    c c                     % COCO-OOD Corruption
}
\toprule

\multirow{2}{*}{\textbf{Model Variant}} &
\multirow{2}{*}{\textbf{Model}} &
\multirow{2}{*}{\makecell[c]{\textbf{Pre-trained}\\\textbf{Backbone}}} &
\multirow{2}{*}{\textbf{Params}} &
\multicolumn{2}{c!{\vrule width .6pt}}{\makecell[c]{\textbf{COCO}}} &
\multicolumn{2}{c!{\vrule width .6pt}}{\makecell[c]{\textbf{HumanArt}}} &
\multicolumn{2}{c!{\vrule width .6pt}}{\makecell[c]{\textbf{COCO-OOD}\\\textbf{Monet}}} &
\multicolumn{2}{c!{\vrule width .6pt}}{\makecell[c]{\textbf{COCO-OOD}\\\textbf{Ukiyo-e}}} &
\multicolumn{2}{c}{\makecell[c]{\textbf{COCO-OOD}\\\textbf{Corruption}}} \\
& & & &
\textbf{AP} & \textbf{AR} &
\textbf{AP} & \textbf{AR} &
\textbf{AP} & \textbf{AR} &
\textbf{AP} & \textbf{AR} &
\textbf{AP} & \textbf{AR} \\
\midrule

% ===================== BODY =====================
\multirow{4}{*}{\textbf{Body}}

& Sapiens-1B~\cite{khirodkar2024sapiens}
& Sapiens ViT
& 1.169B
& 82.1 & 85.9
& 64.3 & 67.4
& 58.8 & 63.3
& 61.5 & 66.2
& 68.6 & 73.2 \\

& Sapiens-2B~\cite{khirodkar2024sapiens}
& Sapiens ViT
& 2.163B
& \textbf{82.2} & \textbf{86.0}
& 69.6 & 72.2
& 59.6 & 64.0
& 62.3 & 66.8
& \textbf{70.3} & 74.8 \\

& GenLoc~\cite{wang2025generalizable}
& Stable Diffusion-v1.5
& 0.98B
& 77.6 & 80.7
& 67.0 & 70.8
& N/A & N/A
& N/A & N/A
& N/A & N/A \\

& \textbf{SDPose (Ours)}
& Stable Diffusion-v2
& 0.98B
& 81.2 & 85.3
& \textbf{71.8} & \textbf{74.6}
& \textbf{64.0} & \textbf{68.7}
& \textbf{66.1} & \textbf{70.9}
& 70.1 & \textbf{75.0} \\

\midrule

% ===================== WHOLEBODY =====================
\multirow{3}{*}{\textbf{Wholebody}}

& Sapiens-1B~\cite{khirodkar2024sapiens}
& Sapiens ViT
& 1.169B
& 72.7 & 79.2
& N/A & N/A
& 38.7 & 46.8
& 40.5 & 49.4
& 48.4 & 57.8 \\

& Sapiens-2B~\cite{khirodkar2024sapiens}
& Sapiens ViT
& 2.163B
& \textbf{74.4} & \textbf{81.0}
& N/A & N/A
& 44.4 & 53.0
& 46.6 & 55.8
& 52.8 & 62.5 \\

& \textbf{SDPose (Ours)}
& Stable Diffusion-v2
& 0.98B
& 72.8 & 79.4
& N/A & N/A
& \textbf{48.4} & \textbf{56.0}
& \textbf{50.0} & \textbf{58.2}
& \textbf{54.3} & \textbf{63.1} \\

\bottomrule
\end{tabular}
\end{adjustbox}

\end{table*}

\section{Experiments}
\label{sec:experiments}
\subsection{Experiment Settings}
\subsubsection{Implementation Details}
We fine-tune SDPose from a Stable Diffusion v2 backbone~\cite{rombach2022high}, with text conditioning disabled. The diffusion timestep is fixed to \(t=1000\). We optimize using AdamW~\cite{loshchilov2017decoupled} with a learning rate of \(3\times10^{-5}\). Inputs are resized to \(1024\times768\) and augmented with standard top-down augmentations. Ablations are conducted on 8 NVIDIA A100 (NVLink) GPUs with a total batch size of 128 and no gradient accumulation. The 17-keypoint model is trained for 40 epochs, and the 133-keypoint model for 42 epochs. Our final feature-aggregation variant is trained on 4 NVIDIA H200 GPUs with the same total batch size (128), without gradient accumulation, and with identical input resolution and augmentations; the 17-keypoint and 133-keypoint models are trained for 40 and 42 epochs, respectively.

\subsubsection{Training Datasets}
We train two variants, SDPose Body (17-keypoints) and SDPose Wholebody (133-keypoints), on MS COCO~\cite{lin2014microsoft} and COCO-WholeBody~\cite{jin2020whole}, respectively. All images are processed using standard top-down augmentations, with the input resolution set to 1024 $\times$ 768. Further details are provided in the supplementary materials.

\subsubsection{COCO-OOD} 
\begin{figure}[!t]
    \centering
    \includegraphics[width=0.65\linewidth,height=0.30\textheight,keepaspectratio]{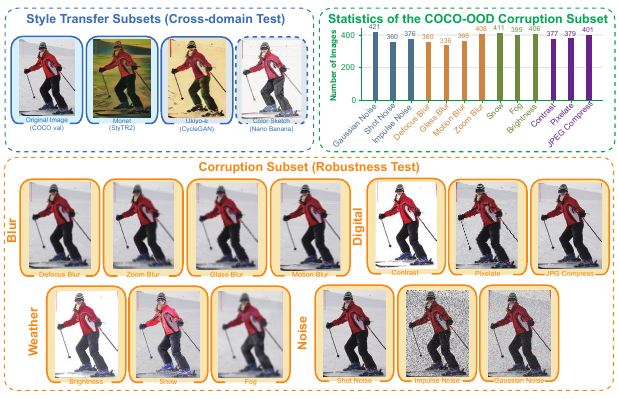}
    \caption{\textbf{Illustration of the COCO-OOD dataset.}}
    \label{fig:coco_ood}
\end{figure}

To complement HumanArt and enable OOD evaluation with matched content and labels, 
we translate all COCO val images into three artistic domains (Fig~\ref{fig:coco_ood}): 
Monet-style paintings using the official StyTR2 framework~\cite{deng2021stytr2}, 
ukiyo-e style using the official CycleGAN implementation~\cite{CycleGAN2017}, 
and color-sketch style using Nano Banana~\cite{nano_banana}. 
For all stylized images, we reuse the original COCO val annotations (bounding boxes and keypoints). We additionally construct a corruption split by applying common low-level corruptions inspired by ImageNet-C~\cite{hendrycks2019benchmarking}. Concretely, we consider 13 corruption types spanning noise (Gaussian, shot, impulse), blur (defocus, glass, motion, zoom), weather (snow, fog, brightness), and digital distortions (contrast, pixelation, JPEG compression). For each image, we uniformly sample one corruption type from this pool and apply it at a fixed severity level (severity=3 in our experiments). The pipeline is implemented with \texttt{albumentations}~\cite{info11020125} (with a custom implementation for impulse noise) and is fully reproducible under a fixed random seed. All images reuse the original COCO val annotations (bounding boxes and keypoints). Please refer to the supplementary materials for details.

\subsubsection{Validation Datasets and Metrics} 
(1) For the Body variant, we evaluate SDPose on MS COCO~\cite{lin2014microsoft} for real-world images, and on HumanArt~\cite{ju2023human} and COCO-OOD for cross-domain benchmarks.
(2) For the Wholebody variant, we evaluate SDPose on COCO-WholeBody~\cite{jin2020whole} and the extended COCO-OOD. Further details of the evaluation datasets and metrics are provided in the supplementary materials.

\subsection{Multi-level Analysis of SD U-Net Features}
\label{sec:single-level-prob}

We analyze SD U-Net features from different upsampling blocks by combining (i) single-level performance evaluation and (ii) representation analysis.
We denote features from the last four upsampling blocks as ${F_1,F_2,F_3,F_4}$.
For single-level evaluation, we train an ablation variant that feeds only one feature level into the heatmap decoder, without feature aggregation, to isolate the contribution of that level.

\noindent\textbf{Single-level performance evaluation.}
As shown in Table~\ref{tab:merged-ablation}, for the 17-keypoint body task, $F_1$ yields the strongest in-domain accuracy (81.3 AP / 85.2 AR on COCO), indicating that the last-block feature preserves the most localization-friendly spatial detail for heatmap decoding.
Under pure style shift (COCO-OOD Monet), the best performance shifts to $F_2$ (64.3 AP / 68.9 AR), while $F_1$ and $F_3$ are comparable (63.5 AP / 68.2 AR).
On HumanArt, the ranking is consistent with COCO: $F_1$ performs best among $\{F_1,F_2,F_3\}$, whereas $F_4$ drops substantially, suggesting that overly coarse features are detrimental to precise keypoint localization.
The same trend is more evident for the wholebody task: $F_2$ outperforms $F_1$ and $F_3$ on COCO-OOD Monet (46.6 AP / 54.8 AR), while $F_4$ degrades sharply on both COCO and COCO-OOD Monet.
Overall, these results reveal a trade-off across upsampling levels: moderately deeper features can improve robustness to style shifts, but overly deep, low-resolution features lose fine-grained spatial cues required for accurate localization.

\noindent\textbf{Representation analysis.}
To understand why robustness varies across levels, we analyze the embedding-space structure of features from pre-trained Sapiens-1B and the SD-v2 U-Net, as well as their pose-finetuned counterparts (Sapiens-Pose and SDPose).
We use t-SNE~\cite{maaten2008visualizing} and silhouette scores~\cite{rousseeuw1987silhouettes} on 1200 person crops (300 COCO instances with three COCO-OOD style variants).
As shown in Fig.~\ref{fig:latent_space}, pre-trained Sapiens exhibits strong style-driven clustering (style silhouette $=0.3469$) and negative instance coherence ($-0.1608$), indicating style-dominant features.
In contrast, SD U-Net features show much weaker style separation and increasingly stronger person-instance coherence when moving to deeper upsampling blocks, with near-zero style silhouettes and rising instance silhouettes (up to $\approx 0.22$).
After pose finetuning, SDPose forms tighter instance clusters than Sapiens-Pose, and the mid-level SDPose blocks ($F_2$--$F_3$) achieve the highest instance coherence ($\approx 0.45$--$0.48$) while keeping style silhouettes close to zero.

\noindent\textbf{Implications for feature selection.}
The representation trends align with single-level performance evaluation.
Although $F_1$ and $F_2$ share the same spatial resolution, $F_2$ exhibits more instance-coherent features under style shifts, consistent with its stronger performance on COCO-OOD Monet.
In contrast, going too deep (especially $F_4$) leads to overly coarse representations that hurt keypoint localization.
Therefore, our final design fuses $\{F_1,F_2,F_3\}$: we retain $F_1$ for localization detail, incorporate $F_2/F_3$ for more stable cues under style shifts, and exclude $F_4$ due to its excessive coarseness.

\begin{table*}[t]
\caption{\textbf{Ablation studies on diffusion priors, RGB reconstruction, U-Net feature selection, and representation aggregation.}
All experiments are trained on COCO with 40 epochs (body-17 keypoints) or 
42 epochs (wholebody-133 keypoints).}
\label{tab:merged-ablation}

\centering
\renewcommand{\arraystretch}{1.2}
\setlength{\tabcolsep}{5pt}
\footnotesize

\begin{adjustbox}{max width=\linewidth,center}

\begin{tabular}{
  c!{\vrule width .6pt}   % Model Variant
  c!{\vrule width .6pt}   % Ablation Setting
  c c!{\vrule width .6pt} 
  c c!{\vrule width .6pt}
  c c
}
\toprule

\multirow{2}{*}{\textbf{Model Variant}} &
\multirow{2}{*}{\textbf{Ablation Setting}} &
\multicolumn{2}{c!{\vrule width .6pt}}{\textbf{COCO}} &
\multicolumn{2}{c!{\vrule width .6pt}}{\textbf{HumanArt}} &
\multicolumn{2}{c}{\textbf{COCO-OOD Monet}}
\\

& &
\textbf{AP} & \textbf{AR} &
\textbf{AP} & \textbf{AR} &
\textbf{AP} & \textbf{AR}
\\
\midrule

% ===================== BODY =====================
\multirow{7}{*}{\textbf{SDPose-Body}} 
  & Single Feature Baseline (Last Block $F_1$)
      & \textbf{81.3} & 85.2 & 71.2 & 73.9 & 63.5 & 68.2 \\

& + Feature Aggregation Module
      & 81.2 (-0.1) & \textbf{85.3} (+0.1) & \textbf{71.8} (+0.6) & \textbf{74.6} (+0.7) & \textbf{64.0} (+0.5) & \textbf{68.7} (+0.5) \\

  & w/o RGB Reconstruction
      & 80.8 (-0.5) & 84.9 (-0.3) 
      & 69.8 (-1.4) & 72.6 (-1.3)
      & 62.5 (-1.0) & 67.3 (-0.9) \\

  & w/o Diffusion Priors
      & 74.9 (-6.4) & 79.4 (-5.8)
      & 53.8 (-17.4) & 58.0 (-15.9)
      & 52.7 (-10.8) & 57.9 (-10.3) \\
      
\cmidrule{2-8}

  & Second-to-last Block ($F_2$)
      & 81.1 & 85.0 & 70.4 & 73.3 & \textbf{64.3} & \textbf{68.9} \\

  & Third-to-last Block ($F_3$)
      & 81.0 & 85.1 & 70.6 & 73.3 & 63.5 & 68.2 \\

  & Fourth-to-last Block ($F_4$)
      & 79.2 & 83.4 & 65.0 & 68.1 & 58.1 & 62.8 \\

\midrule

% ===================== WHOLEBODY =====================
\multirow{4}{*}{\textbf{SDPose-Wholebody}} 
  
  & Single Feature Baseline (Last Block $F_1$)
      & 70.5 & 77.5 & N/A & N/A & 44.7 & 53.0 \\

  & + Feature Aggregation Module
  & \textbf{72.8} (+2.3) & \textbf{79.4} (+1.9) & N/A & N/A & \textbf{48.4} (+3.7)& \textbf{56.0} (+3.0)\\

  & Second-to-last Block ($F_2$)
      & 71.5 & 78.4 & N/A & N/A & 46.6 & 54.8 \\

  & Third-to-last Block ($F_3$)
      & 70.4 & 77.6 & N/A & N/A & 45.5 & 53.8 \\

  & Fourth-to-last Block ($F_4$)
      & 64.6 & 72.1 & N/A & N/A & 37.1 & 45.4 \\

\bottomrule
\end{tabular}
\end{adjustbox}
\end{table*}

\subsection{Feature Aggregation}
\label{sec:fusion-design}

We instantiate feature aggregation with a CSPNeXt-style PAFPN~\cite{lyu2022rtmdet} that fuses $\{F_1,F_2,F_3\}$ before the heatmap decoder (see supplementary). The key effect of this design is to combine complementary cues across U-Net levels: deeper features provide more domain-stable semantics and global body structure, while shallower features preserve fine-grained details that are critical for precise localization. This multi-scale coupling improves OOD behavior without trading off in-domain accuracy for SDPose-Body (e.g., +0.6 AP / +0.7 AR on HumanArt with essentially unchanged COCO in Table~\ref{tab:merged-ablation}). The benefit is substantially larger for SDPose-WholeBody, where fine-grained keypoints (hands/face) demand both global context and local detail; accordingly, aggregation yields clear gains on COCO (+2.3 AP / +1.9 AR) and COCO-OOD Monet (+3.7 AP / +3.0 AR). Overall, the fused representation consistently outperforms any single-level feature ($F_2/F_3/F_4$) alone, supporting the hypothesis that cross-level complementarity is necessary for robust whole-body estimation. Overall, feature fusion becomes more necessary when either the output space is more fine-grained (more keypoints) or the input distribution deviates from the training domain.

\subsection{Quantitative and Qualitative Comparison on Real-World Scenes}
\label{sec:exp-quantitative}
For the Body variant, SDPose achieves 81.2 AP / 85.3 AR on the COCO validation set (Table \ref{tab:quantitative}) with only 40 training epochs using a SD-v2 backbone. It matches the accuracy of Sapiens-1B/2B (82.1--82.2 AP) despite a smaller backbone, and surpasses GenLoc (+3.6 AP, +4.6 AR). Figure \ref{fig:real_world} further illustrates robustness on real-world photos, where SDPose rivals Sapiens and corrects its failure cases (e.g., Sichuan opera eye keypoints). For the Wholebody variant, SDPose achieves competitive performance with Sapiens-1B/2B. Further details are provided in the supplementary materials.

\begin{figure}[!t]
    \centering
    \includegraphics[width=0.65\linewidth,height=0.30\textheight,keepaspectratio]{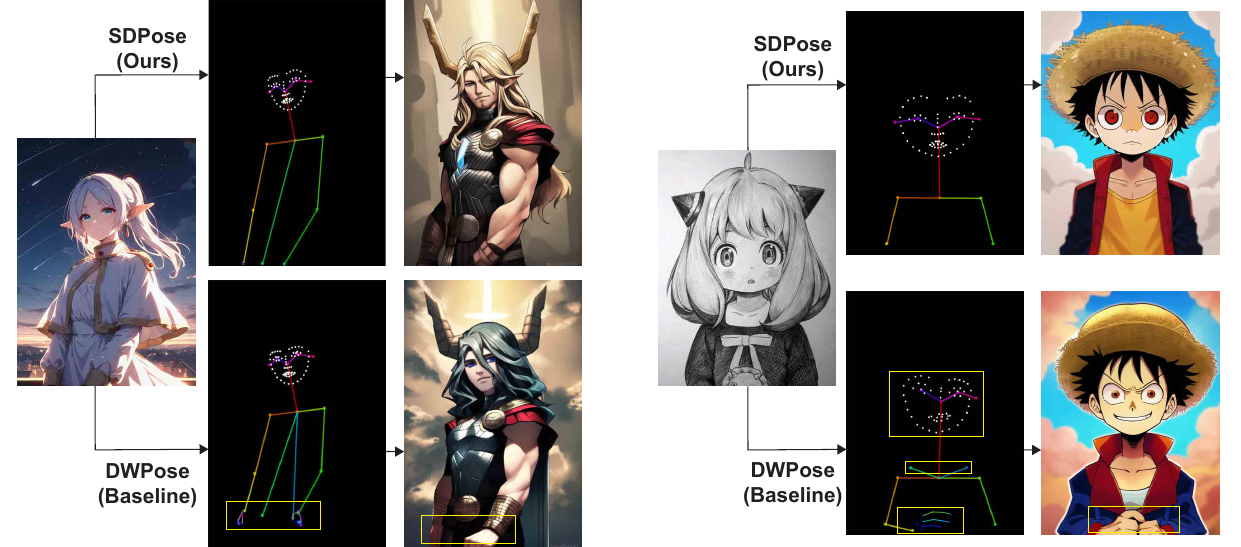}
\caption{\textbf{Visualization of pose-guided image generation results.} The lower images illustrate results from the baseline, which combines a pre-trained ControlNet with the DWPose estimator. In comparison, the upper images show results obtained using our SDPose as the pose annotator. Yellow boxes highlight baseline failures. Prompts, random seeds, and other settings are kept identical for fairness.}
    \label{fig:controlnet}
\end{figure}

\subsection{SDPose's Strong OOD Robustness}

In this section, we demonstrate the OOD robustness of SDPose using  quantitative evaluation on HumanArt and our COCO-OOD benchmark. As shown in Table~\ref{tab:quantitative}, SDPose achieves strong zero-shot results on HumanArt and COCO-OOD with fewer training epochs and a smaller parameter budget. On COCO-OOD WholeBody, SDPose continues to demonstrate strong out-of-domain robustness. As shown in Fig.~\ref{fig:teaser}, SDPose achieves more accurate body pose estimation across diverse animation styles and humanoid robots compared with baseline models. Additional qualitative results on whole-body pose estimation in stylized paintings are provided in the supplementary materials.

\subsection{Ablations on Auxiliary Reconstruction and Diffusion Priors}

We also conduct ablations to validate our designs. For the “w/o diffusion priors” variant, we train the U-Net from scratch (no pre-trained priors). For the “w/o RGB recon.” variant, we disable only the auxiliary RGB reconstruction branch; all other settings remain identical.
From Table~\ref{tab:merged-ablation}, two trends emerge. First, removing the RGB branch yields a consistent but modest AP/AR drop on COCO that becomes more pronounced on HumanArt and COCO-OOD, suggesting that the auxiliary reconstruction acts as a useful regularizer and improves generalization under domain shift. Second, removing diffusion priors causes a much larger degradation, especially on the OOD benchmarks, highlighting that the pre-trained generative priors are the primary source of SDPose’s generalization.

\section{Downstream Applications}

\subsection{Better Pose-guided Image Generation}

For human or humanoid character generation, an accurate skeleton is essential for transferring poses between characters. Traditional pose estimators often fail to precisely capture the skeletons of art-based human or humanoid characters. Our method provides a generalizable pose estimation approach that can benefit animation production. As shown in Fig.~\ref{fig:controlnet}, we compare ControlNet~\cite{zhang2023adding} outputs using DWPose as the baseline pose annotator. Notably, our SDPose yields more precise and detailed skeletons than DWPose, enabling reliable pose transfer and high-quality image generation for artistic characters.

\subsection{Pose-guided Video Generation}

\begin{figure}[!t]
    \centering
    \includegraphics[width=0.65\linewidth,height=0.30\textheight,keepaspectratio]{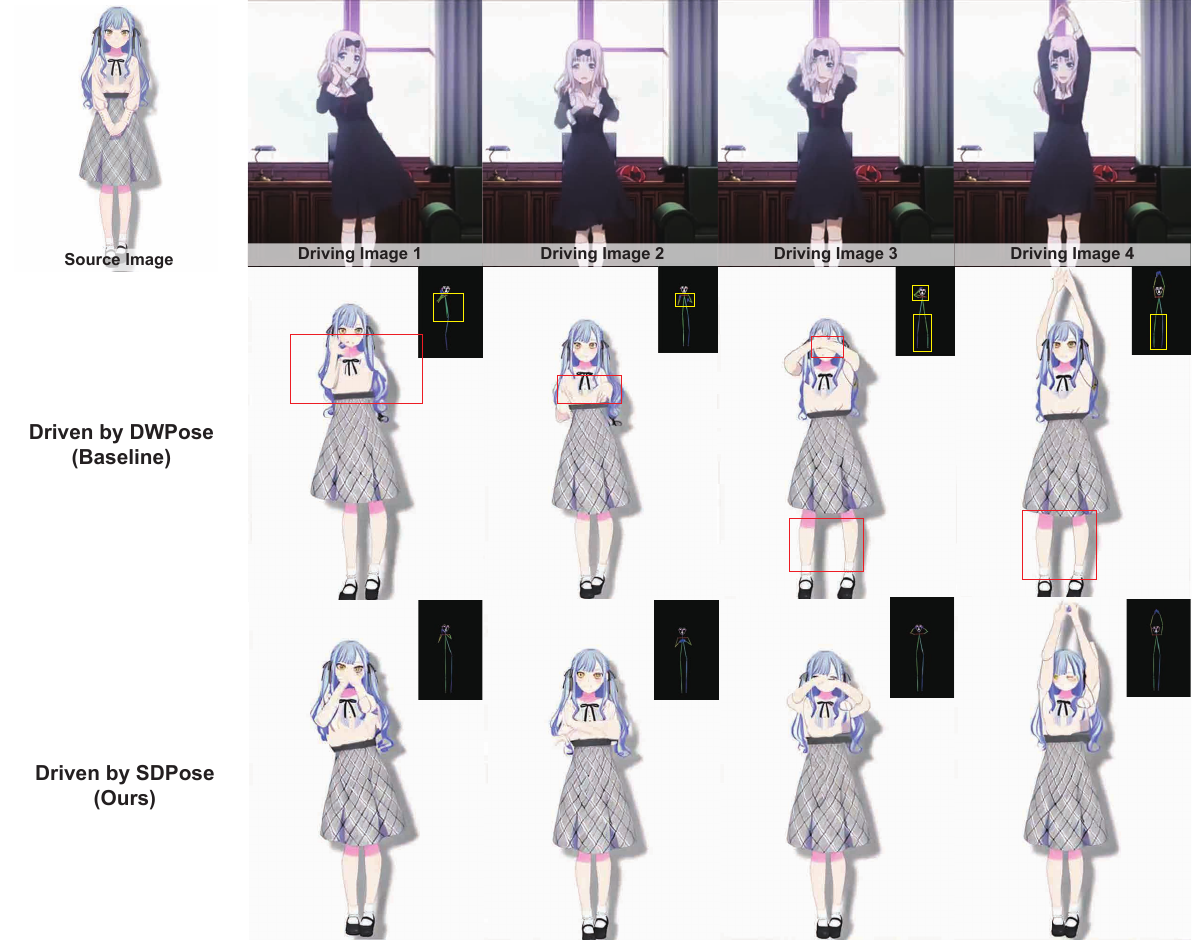}
\caption{\textbf{Qualitative comparison for pose-controlled video generation in the wild. } The first row shows the source image and frames from the driving video. 
The second row shows output video frames generated from the pose sequence estimated by the baseline model DWPose, 
while the third row shows the results guided by our SDPose. 
Red boxes highlight failures in the generated video, and yellow boxes highlight errors in pose estimation.}
    \label{fig:video}
\end{figure}

Recent advances in controlled video generation have gained significant traction~\cite{hu2024animate, guo2023animatediff, kim2024tcan}. 
Despite the progress of video generation models in producing higher-quality outputs, 
extracting reliable control conditions remains critical for high-quality results. 
As shown in Fig.~\ref{fig:video}, SDPose provides more accurate poses for the driving frames, 
enabling reliable pose-sequence transfer from animations to animations. Video frames are generated by Moore-Animated Anyone\footnote{\url{https://github.com/MooreThreads/Moore-AnimateAnyone}}.

\section{Conclusion}

We present \textbf{SDPose}, a pose estimation framework that repurposes a pre-trained Stable Diffusion U-Net as the backbone and predicts keypoints from its upsampling-block features. We further introduce \textbf{COCO-OOD}, a COCO-based benchmark with four subsets, three style-transferred splits for domain-shift evaluation and one corruption split (noise, weather, digital artifacts, and blur) for robustness testing. Across COCO and COCO-WholeBody, SDPose achieves in-domain performance comparable to Sapiens-1B/2B with a smaller parameter scale and shorter fine-tuning schedules. Under domain shifts on HumanArt and COCO-OOD, as well as under common corruptions, SDPose achieves the best results among compared methods under the same evaluation protocol, establishing strong zero-shot performance on HumanArt and multiple COCO-OOD subsets.

\clearpage
{
    \small
    \bibliographystyle{plainnat}
    \bibliography{main}
}

\clearpage
\appendix

\section*{Supplementary Materials for \textbf{SDPose}: Exploiting Diffusion Priors for Out-of-Domain and Robust Pose Estimation}

\section{Experiment Settings}
\label{appendix:training}
\subsection{Implementation Details}
We train SDPose exclusively on the COCO 2017 person keypoints \emph{train2017} split (no extra data), with text conditioning disabled. The diffusion timestep is fixed to \(t=1000\). We optimize using AdamW with a learning rate of \(3\times10^{-5}\). Inputs are resized to \(1024\times768\) and augmented with standard top-down augmentations. Unless otherwise stated, ablations are conducted on 8 NVIDIA A100 (NVLink) GPUs with a total batch size of 128 and no gradient accumulation. The 17-keypoint model is trained for 40 epochs (about 3 days), and the 133-keypoint model for 42 epochs (about 3.5 days). Our final feature-aggregation variant is trained on 4 NVIDIA H200 GPUs with the same total batch size (128), again without gradient accumulation, and with identical input resolution and augmentations; the 17-keypoint and 133-keypoint models are trained for 40 and 42 epochs, respectively, with comparable wall-clock time.

\subsection{Training Datasets}

\textbf{COCO 2017 Keypoint Detection} We train the 17-keypoint variant on the COCO-2017 person keypoint detection dataset~\cite{lin2014microsoft}. The full COCO release contains more than 200,000 images and about 250,000 person instances. Person keypoint annotations follow the 17-point format (nose, eyes, ears, shoulders, elbows, wrists, hips, knees, and ankles).

\noindent \textbf{COCO Wholebody} To further evaluate large-scale whole-body keypoint estimation, we adopt COCO-WholeBody~\cite{jin2020whole}, an extended benchmark built on top of COCO images. COCO-WholeBody augments the original 17 body joints with fine-grained annotations of foot (6 keypoints), face (68 keypoints), and hands (42 keypoints for hands), resulting in a total of 133 keypoints per person. The dataset provides consistent whole-body annotations across the same training and validation splits as COCO-2017, enabling both fair comparison with standard pose estimation methods and comprehensive evaluation under the whole-body setting.

\subsection{Augmentation Details}
The training pipeline first loads the input image and computes the bounding box center and scale. 
It applies random horizontal flipping, half-body augmentation, and random bounding box transformations. 
The image is then affine-transformed to the target input resolution using UDP~\cite{huang2020devil}. 
Albumentations-based augmentations are then applied, including Gaussian blur (\(p=0.1\)), median blur (\(p=0.1\)), and coarse dropout (\(p=1.0\), with up to one hole of size \(20\%\)–\(40\%\) of the image).

\subsection{Evaluation Datasets and Metrics}

\textbf{Evaluation Datasets}

\noindent \textbf{COCO 2017 Keypoint Detection.}
For in-domain evaluation, we use the COCO-2017 validation set~\cite{lin2014microsoft} annotated with 17 body keypoints, bounding boxes, and visibility flags. Following the standard top-down evaluation protocol, we generate person crops from the COCO-released detection results (\texttt{COCO\_val201\allowbreak7\_detections\_AP\_H\_70\_person.json}), and report COCO keypoint AP/AR on this diverse in-the-wild dataset.

\noindent \textbf{HumanArt.}
We use HumanArt~\cite{ju2023human} as an cross domain benchmark: 50k human centric images across 20 scenarios (5 natural, 15 artistic: oil painting, sculpture, cartoon, sketch, stained glass, Ukiyo-e, watercolor, etc.) with annotations for boxes and 2D keypoints. 
We follow the official protocol and report keypoint AP/AR to assess robustness under artistic domain shift.

\noindent \textbf{COCO-WholeBody (133-keypoint whole-body).}
We train and evaluate a 133-keypoint variant on COCO-WholeBody~\cite{jin2020whole}, which shares COCO’s train/val split.
Each person has 133 keypoints (17 body, 6 foot, 68 face, 42 hand) plus boxes for person/face/left/right hand.
We follow the official protocol and report Whole-Body AP and part-wise AP (body/foot/face/hand).
This dataset spans diverse in-the-wild scenes and stresses fine-grained articulation, complementing COCO for structured keypoint evaluation.

\noindent \textbf{Metrics}
We follow the standard COCO keypoint evaluation protocol, which is based on the Object Keypoint Similarity (OKS). 
For each keypoint $i$, the similarity is defined as
\[
KS_i = \exp\!\left(-\frac{d_i^2}{2 s^2 k_i^2}\right),
\]
where $d_i$ denotes the Euclidean distance between predicted and ground-truth keypoints, $s$ is the object scale (square root of the segmentation area), and $k_i$ is a per-keypoint constant controlling falloff. 
The OKS for an instance is the average $KS_i$ over visible keypoints:
\[
OKS = \frac{\sum_i KS_i \cdot \delta(v_i > 0)}{\sum_i \delta(v_i > 0)},
\]
where $v_i$ is the visibility flag. 
Using OKS as the matching criterion, COCO computes Average Precision (AP) as the mean precision over OKS thresholds $[0.50:0.05:0.95]$, and Average Recall (AR) analogously as the mean recall across the same thresholds.

\subsection{Details of COCO-OOD}

\begin{figure*}[h]
    \centering
    \includegraphics[width=0.8\linewidth]{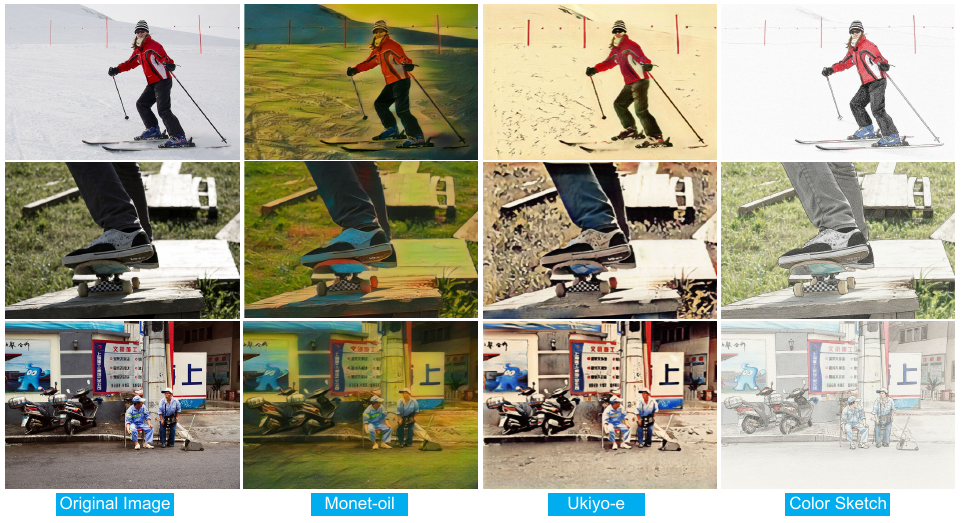}
    \caption{
        \textbf{Visual illustration of the style-transferred subsets in the COCO-OOD benchmark.} We create an OOD split of the COCO validation set by stylizing natural images into three distinct artistic domains: Monet (via StyTR2), Ukiyo-e (via CycleGAN), and Color-sketch (via Nano Banana), enabling comprehensive evaluation under diverse style shifts.
    }
    \label{COCO-OOD Full}
\end{figure*}

To complement HumanArt and enable cross domain and robustness evaluation under matched content and annotations, we construct COCO-OOD (Fig.~\ref{COCO-OOD Full}) by applying a diverse set of image transformations to the original COCO val2017 images while preserving all human bounding boxes and keypoint annotations. Specifically, COCO-OOD consists of four distinct subsets designed to evaluate two primary types of distribution shifts: three style-transferred splits for artistic domain shifts, and one corruption split for low-level robustness. This comprehensive design allows us to assess both cross-domain generalization (via stylistic changes) and algorithmic resilience to common image degradations within a unified benchmark.

\noindent\textbf{Artistic Stylization (COCO-OOD-Style).} The first three subsets target domain-level appearance shifts via style transfer. Importantly, for fair comparison and to avoid introducing priors from large-scale pre-trained diffusion models, we intentionally adopt the earlier CycleGAN~\cite{CycleGAN2017} and StyTr2~\cite{deng2021stytr2} frameworks rather than more recent style transfer approaches. For the \textit{Monet} oil-painting variant, we employ StyTr2, which produces high-fidelity oil-painting textures and color palettes while maintaining overall scene geometry. For the \textit{Ukiyo-e} variant, we adopt the official CycleGAN implementation to translate natural photographs into the Ukiyo-e domain. Finally, we include a \textit{Color Sketch} variant generated using the Nano-Banana model~\cite{nano_banana}. While Nano-Banana preserves global shape in most cases, its stylization can occasionally introduce slight pixel-level spatial misalignment with respect to the source images; we therefore use this variant only as a supplemental resource for latent-space t-SNE analysis and exclude it from quantitative evaluation. By drawing from multiple style-transfer frameworks, COCO-OOD-Style reduces the bias introduced by any single model and produces stylistic shifts with diverse visual characteristics.

\noindent\textbf{Corruption (COCO-OOD-Corruption).}
We additionally build a corruption split on COCO \texttt{val2017} by applying ImageNet-C~\cite{hendrycks2019benchmarking} inspired low-level corruptions.
We consider 13 corruption types spanning noise (Gaussian, shot, impulse), blur (defocus, glass, motion, zoom), weather/illumination (snow, fog, brightness), and digital distortions (contrast, pixelation, JPEG compression).
For each image, we uniformly sample one corruption type from this pool and apply it at a fixed severity level (severity=3 in all experiments).
The pipeline is implemented with \texttt{albumentations}, with a custom implementation for impulse noise.
We fix the random seed for reproducibility, and we reuse the original COCO \texttt{val2017} annotations (bounding boxes and keypoints) since the corruptions do not change spatial geometry.

\subsection{Details of Feature Aggregation Module}

\begin{figure*}[!t]
    \centering
    \includegraphics[width=1.0\linewidth]{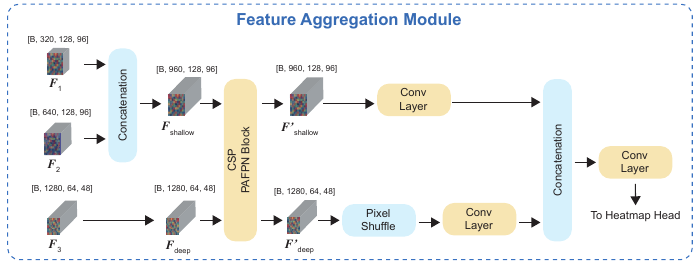}
    \caption{
        \textbf{Hierarchical features are fused through a CSP PAFPN block. Deep features are upsampled via PixelShuffle to align with shallow features for final heatmap prediction.} }
    \label{Fusion}
\end{figure*}

As shown in Fig.~\ref{Fusion}, we aggregate $\{F_1,F_2,F_3\}$ with a CSP-PAFPN-style block~\cite{lyu2022rtmdet}.
We first build a shallow branch by concatenating $(F_1,F_2)$ and a deep branch from $F_3$, and perform bidirectional feature fusion within PAFPN (top-down and bottom-up) to propagate semantics to high-resolution features and feed back fine details to deeper levels.
After PAFPN fusion, we upsample the deep branch via PixelShuffle to match the spatial resolution of the shallow branch, and further refine both branches with lightweight convolution layers for channel mixing.
Finally, the aligned features are concatenated and projected by a convolution layer to obtain the fused representation for the heatmap head.

\section{Details of Quantitative Comparison and Additional Qualitative Comparison}

\subsection{Full Quantitative Comparison on COCO}
SDPose achieves 81.2 AP / 85.3 AR on the COCO validation set (Table \ref{tab:coco-val-full}) with only 40 training epochs using a 0.95B SD-v2 backbone. It matches the accuracy of Sapiens (82.1--82.2 AP) despite requiring 5× fewer epochs and a smaller backbone. SDPose also outperforms ViTPose++ (+1.8 AP), which relies on multiple auxiliary datasets, while being trained solely on COCO.
\begin{table}[!t]
\caption{\textbf{Quantitative comparison on the COCO-2017 validation set.}
All results are self-reported by the corresponding works.}
\label{tab:coco-val-full}
\centering
\renewcommand{\arraystretch}{1.2}
\begin{tabular}{c c c c}
\noalign{\hrule height 1.2pt}
\textbf{Model} & \textbf{Input Size} & \textbf{AP} & \textbf{AR} \\
\hline
SimpleBaseline~\cite{xiao2018simple} & 256$\times$192 & 73.5 & 79.0 \\
HRNet~\cite{sun2019deep}             & 384$\times$288 & 76.3 & 81.2 \\
HRFormer~\cite{YuanFHLZCW21}         & 256$\times$192 & 77.2 & 82.0 \\
ViTPose-S~\cite{xu2022vitpose}       & 256$\times$192 & 73.8 & 79.2 \\
ViTPose-B~\cite{xu2022vitpose}       & 256$\times$192 & 75.8 & 81.1 \\
ViTPose-L~\cite{xu2022vitpose}       & 256$\times$192 & 78.3 & 83.5 \\
ViTPose-H~\cite{xu2022vitpose}       & 256$\times$192 & 79.1 & 84.1 \\
ViTPose++-S~\cite{xu2023vitpose++}   & 256$\times$192 & 75.8 & 81.0 \\
ViTPose++-B~\cite{xu2023vitpose++}   & 256$\times$192 & 77.0 & 82.6 \\
ViTPose++-L~\cite{xu2023vitpose++}   & 256$\times$192 & 78.6 & 84.1 \\
ViTPose++-H~\cite{xu2023vitpose++}   & 256$\times$192 & 79.4 & 84.8 \\
Sapiens-0.3B~\cite{khirodkar2024sapiens} & 1024$\times$768 & 79.6 & 83.6 \\
Sapiens-0.6B~\cite{khirodkar2024sapiens} & 1024$\times$768 & 81.2 & 84.9 \\
Sapiens-1B~\cite{khirodkar2024sapiens}   & 1024$\times$768 & 82.1 & 85.9 \\
\textbf{Sapiens-2B}~\cite{khirodkar2024sapiens}
& 1024$\times$768 & \textbf{82.2} & \textbf{86.0} \\
ProbPose~\cite{purkrabek2025probpose} & N/A & 76.6 & 76.4 \\
\hline
Painter~\cite{wang_images_2023}                       & N/A & 72.5 & - \\
InstructDiffusion~\cite{geng_instructdiffusion_2024}  & N/A & 71.2 & - \\
X-Pose-T~\cite{leonardis_x-pose_2025}                 & N/A & 74.2 & - \\
RTMO-L~\cite{lu_rtmo_nodate}                          & N/A & 72.7 & 76.9 \\
Group Pose~\cite{liu_group_2023}                       & N/A & 74.8 & - \\
ED-Pose~\cite{yang_explicit_2023}                     & N/A & 75.8 & - \\
\hline
\textbf{SDPose (Ours)} & 1024$\times$768 & 81.2 & 85.3 \\
\noalign{\hrule height 1.2pt}
\end{tabular}
\end{table}

\begin{table}[!t]
\caption{\textbf{Quantitative comparison on the HumanArt dataset.}
All results are self-reported by the corresponding works except *, whose value is the reproduction reported by InstructDiffusion~\cite{geng_instructdiffusion_2024}. X-Pose is trained on HumanArt and is therefore excluded from the zero-shot evaluation.}
\label{tab:humanart-val}
\centering
\renewcommand{\arraystretch}{1.2}
\begin{adjustbox}{max width=\linewidth,center}

\begin{tabular}{c c c c c c}
\noalign{\hrule height 1.2pt}
\textbf{Model} & \textbf{AP} & \textbf{AP$^{50}$} & \textbf{AP$^{75}$}
& \textbf{AR} & \textbf{AR$^{50}$} \\
\hline
RTMPose-T~\cite{jiang2023rtmpose}  & 44.4 & 72.5 & 45.3 & 48.8 & 75.0 \\
RTMPose-S~\cite{jiang2023rtmpose}  & 48.0 & 73.9 & 49.8 & 52.1 & 76.3 \\
RTMPose-M~\cite{jiang2023rtmpose}  & 53.2 & 76.5 & 56.3 & 57.1 & 78.9 \\
RTMPose-L~\cite{jiang2023rtmpose}  & 56.4 & 78.9 & 60.2 & 59.9 & 80.8 \\
ViTPose-S~\cite{xu2022vitpose}     & 50.7 & 75.8 & 53.1 & 55.1 & 78.0 \\
ViTPose-B~\cite{xu2022vitpose}     & 55.5 & 78.2 & 59.0 & 59.9 & 80.9 \\
ViTPose-L~\cite{xu2022vitpose}     & 63.7 & 83.8 & 68.9 & 67.7 & 85.9 \\
ViTPose-H~\cite{xu2022vitpose}     & 66.5 & 86.0 & 71.5 & 70.1 & 87.1 \\
HRNet-W32~\cite{sun2019deep}       & 53.3 & 77.1 & 56.2 & 57.4 & 79.2 \\
HRNet-W48~\cite{sun2019deep}       & 55.7 & 78.2 & 59.3 & 59.5 & 80.4 \\
Sapiens-1B~\cite{khirodkar2024sapiens}
& 64.3 & 82.1 & 67.9 & 67.4 & 83.7 \\
Sapiens-2B~\cite{khirodkar2024sapiens}
& 69.6 & 85.3 & 73.3 & 72.2 & 86.8 \\
\hline
Painter~\cite{wang_images_2023}
& 12.4* & - & - & - & - \\
InstructDiffusion~\cite{geng_instructdiffusion_2024}
& 51.4 & - & - & - & - \\
X-Pose-T~\cite{leonardis_x-pose_2025}
& \textbf{75.9} & - & - & - & - \\
\hline
\textbf{SDPose (Ours)}
& 71.8 & \textbf{87.3} & \textbf{76.3} & \textbf{74.6} & \textbf{88.6} \\
\noalign{\hrule height 1.2pt}
\end{tabular}
\end{adjustbox}
\end{table}

On HumanArt (Table~\ref{tab:humanart-val}), SDPose sets the best zero-shot performance with 71.8 AP / 74.6 AR, surpassing large-scale foundation baselines under the same COCO-only training: +2.2 AP over Sapiens-2B (69.6 AP) and +5.3 AP over ViTPose-H (66.5 AP), with consistent AR gains. Compared with traditional baselines such as RTMPose~\cite{jiang2023rtmpose} and HRNet~\cite{sun2019deep}, SDPose further delivers substantial improvements in AP, exceeding them by more than 15 points.

\subsection{Full Quantitative Comparison on COCO Wholebody}

\begin{table}[h]
    \centering
    \caption{\textbf{Quantitative comparison on the COCO-WholeBody validation set.}}
    \label{tab:coco-wholebody}
    \renewcommand{\arraystretch}{1.2}
    
    % 如果不需要脚注可注释掉 threeparttable
    % \begin{threeparttable}
    \footnotesize

    % ---- 第一个子表 ----
    \begin{subtable}{\columnwidth}
        \centering
        \caption{\textbf{Body, Feet, Face}}
        \setlength{\tabcolsep}{4pt}
 \begin{adjustbox}{max width=\linewidth,center}

            \begin{tabular}{l c c c c c c}
                \noalign{\hrule height 1.2pt}
                \textbf{Model} & \textbf{Body AP} & \textbf{Body AR} & \textbf{Feet AP} & \textbf{Feet AR} & \textbf{Face AP} & \textbf{Face AR} \\
                \hline
                HRNet~\cite{sun2019deep}         & 70.1 & 77.3 & 58.6 & 69.2 & 72.7 & 78.3 \\
                VitPose$+$-L~\cite{xu2023vitpose++} & 75.3 &  -   & 77.1 &  -   & 63.0 &  -   \\
                VitPose$+$-H~\cite{xu2023vitpose++} & 75.9 &  -   & 77.9 &  -   & 63.6 &  -   \\
                RTMPose-x~\cite{jiang2023rtmpose}   & 71.4 & 78.4 & 69.2 & 81.0 & 88.8 & 92.2 \\
                DWPose-l~\cite{yang2023effective}   & 72.2 & 78.9 & 70.4 & 81.7 & 88.7 & 92.1 \\
                Sapiens-0.3B~\cite{khirodkar2024sapiens} & 66.4 & 73.4 & 67.3 & 78.4 & 87.1 & 91.2 \\
                Sapiens-0.6B~\cite{khirodkar2024sapiens} & 74.3 & 80.2 & 79.4 & 87.0 & 89.5 & 92.9 \\
                Sapiens-1B~\cite{khirodkar2024sapiens}   & 77.4 & 82.9 & 83.0 & 89.8 & 90.7 & 93.6 \\
                Sapiens-2B~\cite{khirodkar2024sapiens}   & \textbf{79.2} & \textbf{84.6} & \textbf{84.1} & \textbf{90.9} & \textbf{91.2} & \textbf{93.8} \\
                \hline
                SDPose-F2 (Ours)   & 77.9 & 83.4 & 81.5 & 88.7 & 88.5 & 92.2 \\
                SDPose (Ours)   & 78.6 & 84.1 & 82.3 & 89.0 & 89.7 & 92.8 \\
                \noalign{\hrule height 1.2pt}
            \end{tabular}
        \end{adjustbox}
    \end{subtable}

    % ---- 第二个子表 ----
    \begin{subtable}{\columnwidth}
        \centering
        \caption{\textbf{Hands and Whole-body}}
        \setlength{\tabcolsep}{4pt} % 与上表保持统一的 tabcolsep，让 resizebox 自己去计算比例
        
            \begin{tabular}{l c c c c c c}
                \noalign{\hrule height 1.2pt}
                \textbf{Model} & \textbf{Hand AP} & \textbf{Hand AR} & \textbf{Whole AP} & \textbf{Whole AR} \\
                \hline
                HRNet~\cite{sun2019deep}         & 51.6 & 60.4 & 58.6 & 67.4 \\
                VitPose$+$-L~\cite{xu2023vitpose++} & 54.2 &  -   & 60.6 &  -   \\
                VitPose$+$-H~\cite{xu2023vitpose++} & 54.7 &  -   & 61.2 &  -   \\
                RTMPose-x~\cite{jiang2023rtmpose}   & 59.0 & 68.5 & 65.3 & 73.3 \\
                DWPose-l~\cite{yang2023effective}   & 62.1 & 71.0 & 66.5 & 74.3 \\
                Sapiens-0.3B~\cite{khirodkar2024sapiens} & 58.1 & 67.1 & 62.0 & 69.4 \\
                Sapiens-0.6B~\cite{khirodkar2024sapiens} & 65.4 & 74.0 & 69.5 & 76.3 \\
                Sapiens-1B~\cite{khirodkar2024sapiens}   & 69.2 & 77.1 & 72.7 & 79.2 \\
                Sapiens-2B~\cite{khirodkar2024sapiens}   & \textbf{70.4} & \textbf{78.1} & \textbf{74.4} & \textbf{81.0} \\
                \hline
                SDPose-F2 (Ours)   & 65.2 & 74.0 & 71.5 & 78.4 \\
                SDPose (Ours)   & 67.3 & 75.4 & 72.8 & 79.4 \\
                \noalign{\hrule height 1.2pt}
            \end{tabular}
        
    \end{subtable}
    
    % \end{threeparttable}
\end{table}

\begin{table}[!t]
    \centering
    \caption{\textbf{Quantitative Comparison on the COCO-OOD-Monet Wholebody validation set.}}
    \label{tab:coco-ood-wholebody-monet}
    \renewcommand{\arraystretch}{1.2}
    
    % 如果你不需要在表下方加脚注，可以把 threeparttable 注释掉
    % \begin{threeparttable} 
    \footnotesize

    % ---- 第一个子表 ----
    \begin{subtable}{\columnwidth} 
        \centering
        \caption{\textbf{Body, Feet, Face}}
        \setlength{\tabcolsep}{4pt}
         \begin{adjustbox}{max width=\linewidth,center}

            \begin{tabular}{l c c c c c c}
                \noalign{\hrule height 1.2pt}
                \textbf{Model} & \textbf{Body AP} & \textbf{Body AR} & \textbf{Feet AP} & \textbf{Feet AR} & \textbf{Face AP} & \textbf{Face AR} \\
                \hline
                Sapiens-1B & 52.1 & 58.6 & 55.9 & 66.2 & 57.6 & 63.0 \\
                Sapiens-2B & 59.9 & 65.8 & 63.8 & 72.4 & 58.4 & 64.2 \\
                \hline
                SDPose-F2 (Ours) & 60.0 & 66.1 & 62.5 & 72.0 & 64.2 & 69.9 \\
                SDPose (Ours) & \textbf{61.3} & \textbf{67.2} & \textbf{64.3} & \textbf{72.9} & \textbf{66.5} & \textbf{71.3} \\
                \noalign{\hrule height 1.2pt}
            \end{tabular}
        \end{adjustbox}
    \end{subtable}

    % ---- 第二个子表 ----
    \begin{subtable}{\columnwidth}
        \centering
        \caption{\textbf{Hands and Whole-body}}
        \setlength{\tabcolsep}{4pt} % 统一为4pt，让resizebox自动缩放
         \begin{adjustbox}{max width=\linewidth,center}

            \begin{tabular}{l c c c c c c}
                \noalign{\hrule height 1.2pt}
                \textbf{Model} & \textbf{L-Hand AP} & \textbf{L-Hand AR} & \textbf{R-Hand AP} & \textbf{R-Hand AR} & \textbf{Whole AP} & \textbf{Whole AR} \\
                \hline
                Sapiens-1B & 43.1 & 52.0 & 41.5 & 50.6 & 38.7 & 46.8 \\
                Sapiens-2B & 48.2 & 56.8 & 46.8 & 55.3 & 44.4 & 53.0 \\
                \hline
                SDPose-F2 (Ours) & 46.3 & 55.2 & 44.9 & 54.4 & 46.6 & 54.8\\ 
                SDPose (Ours) & \textbf{48.8} & \textbf{57.1} & \textbf{47.9} & \textbf{56.9} & \textbf{48.4} & \textbf{56.0} 
                \\
                \noalign{\hrule height 1.2pt}
            \end{tabular}
        \end{adjustbox}
    \end{subtable}

    % \end{threeparttable}
\end{table}

\begin{table}[!t]
    \centering
    \caption{\textbf{Quantitative Comparison on the COCO-OOD-Ukiyoe Wholebody validation set.}}
    \label{tab:coco-ood-wholebody-ukiyoe}
    \renewcommand{\arraystretch}{1.2}
    
    \footnotesize

    \begin{subtable}{1\columnwidth}
        \centering
        \caption{\textbf{Body, Feet, Face}}
        \setlength{\tabcolsep}{4pt}
        \begin{adjustbox}{max width=\linewidth,center}

            \begin{tabular}{l c c c c c c}
                \noalign{\hrule height 1.2pt}
                \textbf{Model} & \textbf{Body AP} & \textbf{Body AR} & \textbf{Feet AP} & \textbf{Feet AR} & \textbf{Face AP} & \textbf{Face AR} \\
                \hline
                Sapiens-1B & 54.0 & 60.9 & 57.9 & 69.9 & 58.7 & 65.3 \\
                Sapiens-2B & 62.2 & 68.2 & \textbf{67.3} & \textbf{76.5} & 59.9 & 66.9 \\
                \hline
                SDPose (Ours) & \textbf{62.8} & \textbf{68.8} & 66.1 & 75.4 & \textbf{67.4} & \textbf{73.0} \\
                \noalign{\hrule height 1.2pt}
            \end{tabular}
        \end{adjustbox}
    \end{subtable}

    \begin{subtable}{1\columnwidth}
        \centering
        \caption{\textbf{Hands and Whole-body}}
        \setlength{\tabcolsep}{4pt}
        \begin{adjustbox}{max width=\linewidth,center}

            \begin{tabular}{l c c c c c c}
                \noalign{\hrule height 1.2pt}
                \textbf{Model} & \textbf{L-Hand AP} & \textbf{L-Hand AR} & \textbf{R-Hand AP} & \textbf{R-Hand AR} & \textbf{Whole AP} & \textbf{Whole AR} \\
                \hline
                Sapiens-1B & 43.7 & 52.8 & 40.9 & 50.6 & 40.5 & 49.4 \\
                Sapiens-2B & \textbf{49.8} & \textbf{58.7} & 46.7 & 55.7 & 46.6 & 55.8 \\
                \hline
                SDPose (Ours) & 48.8 & 58.0 & \textbf{46.8} & \textbf{56.5} & \textbf{50.0} & \textbf{58.2} \\
                \noalign{\hrule height 1.2pt}
            \end{tabular}
        \end{adjustbox}
    \end{subtable}

\end{table}

\begin{table}[h]
    \centering
    \caption{\textbf{Quantitative Comparison on the COCO-OOD-Corruption Wholebody validation set.}}
    \label{tab:coco-ood-wholebody-corruption}
    \renewcommand{\arraystretch}{1.2}
    
    % 如果你不需要在表下方加脚注，可以把 threeparttable 注释掉
    % \begin{threeparttable} 
    \footnotesize

    % ---- 第一个子表 ----
    \begin{subtable}{1\columnwidth}
        \centering
        \caption{\textbf{Body, Feet, Face}}
        \setlength{\tabcolsep}{4pt}
         \begin{adjustbox}{max width=\linewidth,center}

            \begin{tabular}{l c c c c c c}
                \noalign{\hrule height 1.2pt}
                \textbf{Model} & \textbf{Body AP} & \textbf{Body AR} & \textbf{Feet AP} & \textbf{Feet AR} & \textbf{Face AP} & \textbf{Face AR} \\
                \hline
                Sapiens-1B & 62.3 & 68.7 & 67.1 & 76.4 & 57.0 & 64.7 \\
                Sapiens-2B & \textbf{67.9} & \textbf{73.9} & \textbf{73.7} & \textbf{81.5} & 58.0 & 65.7 \\
                \hline
                SDPose (Ours) & 67.8 & \textbf{73.9} & 71.3 & 79.4 & \textbf{62.8} & \textbf{69.4} \\
                \noalign{\hrule height 1.2pt}
            \end{tabular}
        \end{adjustbox}
    \end{subtable}

    % ---- 第二个子表 ----
    \begin{subtable}{1\columnwidth}
        \centering
        \caption{\textbf{Hands and Whole-body}}
        \setlength{\tabcolsep}{4pt}
         \begin{adjustbox}{max width=\linewidth,center}

            \begin{tabular}{l c c c c c c}
                \noalign{\hrule height 1.2pt}
                \textbf{Model} & \textbf{L-Hand AP} & \textbf{L-Hand AR} & \textbf{R-Hand AP} & \textbf{R-Hand AR} & \textbf{Whole AP} & \textbf{Whole AR} \\
                \hline
                Sapiens-1B & 53.1 & 61.8 & 50.6 & 59.4 & 48.4 & 57.8 \\
                Sapiens-2B & \textbf{57.6} & \textbf{66.0} & \textbf{54.6} & \textbf{63.4} & 52.8 & 62.5 \\
                \hline
                SDPose (Ours) & 55.3 & 63.8 & 53.3 & 62.1 & \textbf{54.3} & \textbf{63.1} \\
                \noalign{\hrule height 1.2pt}
            \end{tabular}
        \end{adjustbox}
    \end{subtable}

    % \end{threeparttable}
\end{table}

As shown in Table~\ref{tab:coco-wholebody}, SDPose achieves 72.8 AP and 79.4 AR on the COCO-WholeBody validation set. This result is highly competitive even when compared to the large-scale Sapiens~\cite{khirodkar2024sapiens} models. While Sapiens-2B reaches 74.4 AP using over 2 billion parameters and an extensive training schedule, SDPose attains comparable accuracy utilizing a smaller backbone trained for only 42 epochs.

In our sub-part analysis, SDPose closely matches Sapiens-2B on body (78.6 vs. 79.2) and feet (82.3 vs. 84.1), while maintaining strong performance on face keypoints (89.7 vs. 91.2). The largest gap is observed in hand keypoints (67.3 vs. 70.4), reflecting the intrinsic difficulty of capturing fine-grained articulations. Nevertheless, when compared with classical baselines such as HRNet~\cite{sun2019deep}, RTMPose~\cite{jiang2023rtmpose}, and DWPose~\cite{yang2023effective}, SDPose demonstrates substantial improvements ranging from +6 to +14 AP in the overall whole-body evaluation.

Furthermore, when compared to the ablation baseline SDPose-F2, which is tuned solely on the second-to-last feature from the SD U-Net, our full SDPose model yields a clear +2.1 AP boost on hand keypoints. This specific gain highlights the necessity of our proposed feature aggregation strategy for accurately detecting fine-grained keypoints.

\subsection{Full Quantitative Comparison on COCO-OOD Wholebody}

Detailed whole-body pose estimation results on COCO-OOD Monet are reported in Table~\ref{tab:coco-ood-wholebody-monet}. Breaking down by body part, SDPose matches or surpasses Sapiens in all regions: body (61.3 AP vs. 59.9), feet (64.3 vs. 63.8), face (66.5 vs. 58.4), left hand (48.8 vs. 48.2), and right hand (47.9 vs. 46.8), highlighting its reliability under appearance shifts. Consistent with the results from COCO Wholebody, the feature aggregation greatly improves the performance of fine-grained detection for feet and hand keypoints.

Detailed whole-body pose estimation results on COCO-OOD Ukiyo-e are presented in Table~\ref{tab:coco-ood-wholebody-ukiyoe}.
Breaking down by region, SDPose slightly outperforms Sapiens-2B on body keypoints (62.8 vs. 62.2 AP), while remaining competitive on feet (66.1 vs. 67.3 AP).
A substantially larger advantage is observed for facial landmarks, where SDPose improves over Sapiens-2B by +7.5 AP (67.4 vs. 59.9), indicating stronger robustness to artistic appearance shifts.
For hand keypoints, SDPose remains close to Sapiens-2B on the left hand (48.8 vs. 49.8 AP) and slightly improves the right hand (46.8 vs. 46.7 AP).
Importantly, SDPose achieves the highest whole-body performance with 50.0 AP / 58.2 AR, outperforming Sapiens-2B by +3.4 AP / +2.4 AR and demonstrating stronger overall robustness under Ukiyo-e style transformations.

Table~\ref{tab:coco-ood-wholebody-corruption} reports the quantitative comparison on the COCO-OOD Corruption dataset. Under various image corruptions, SDPose demonstrates superior robustness, surpassing Sapiens-2B in overall Whole AP (54.3 vs. 52.8). Notably, SDPose exhibits a remarkable advantage in facial keypoint detection, exceeding Sapiens-2B by 4.8 AP (62.8 vs. 58.0). Although Sapiens-2B retains a slight edge in localizing extremities such as hands and feet, SDPose maintains highly competitive performance on the main body (67.8 AP vs. 67.9 AP). These results suggest that while Sapiens may excel at localized limb predictions under noise, SDPose provides a more balanced and structurally robust understanding of the entire human body.

\subsection{Qualitative Comparison for Whole-Body Pose Estimation}

\begin{figure}[h]
    \centering
    \includegraphics[width=0.65\linewidth,height=0.30\textheight,keepaspectratio]{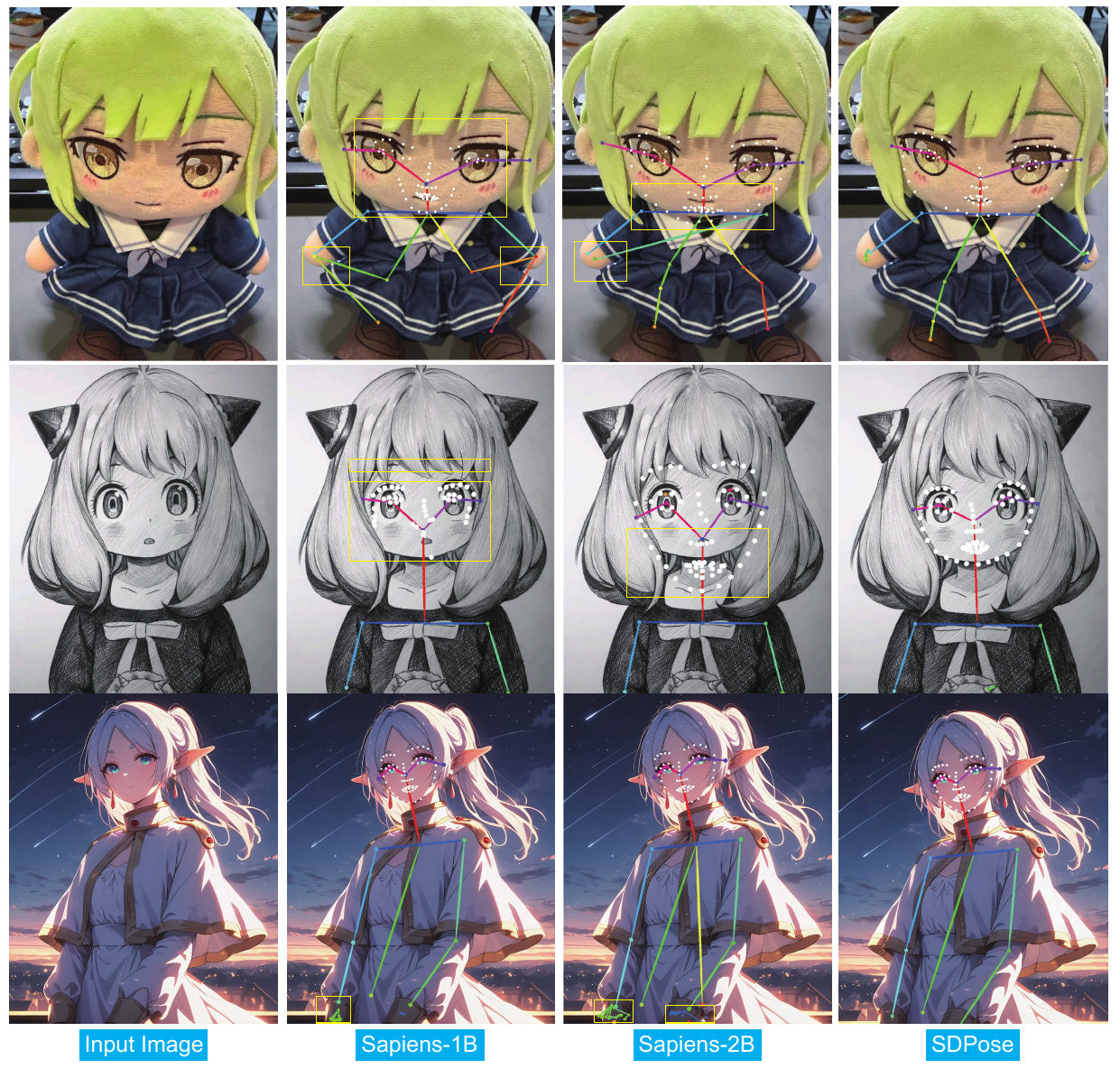}
\caption{\textbf{Comparison on Stylized Paintings: Sapiens WholeBody vs.\ SDPose WholeBody.} All erroneous predictions are highlighted with yellow boxes. SDPose yields fewer false positives and notably better facial keypoint localization.}
\label{fig:ood_stylized}
\end{figure}

{
\subsection{SDv1.5 Baseline}

As shown in Table~\ref{tab:sd_version}, SDPose achieves similar performance with SD-v1.5 and SD-v2 (81.2 vs.\ 81.3 AP). More importantly, using the same SD-v1.5 backbone family as GenLoc, SDPose achieves 81.2 AP / 85.3 AR, outperforming GenLoc by 3.6 AP / 4.6 AR. This controlled comparison shows that the gain over GenLoc cannot be simply attributed to the use of SD-v2.

\begin{table}[!t]
\caption{\textbf{Controlled comparison with Stable Diffusion v1.5 on COCO.}
For the SDPose variants, we use only the last-layer U-Net feature without the feature aggregation module.}
\label{tab:sd_version}
\centering
\renewcommand{\arraystretch}{1.2}

\begin{adjustbox}{max width=\linewidth,center}

\begin{tabular}{c c c c c c c}
\noalign{\hrule height 1.2pt}
\textbf{Model} &
\textbf{Pre-trained Backbone} &
\textbf{Params.} &
\textbf{Train Dataset} &
\textbf{Train Epochs} &
\textbf{AP} &
\textbf{AR} \\
\hline
GenLoc~\cite{wang2025generalizable}
& Stable Diffusion-v1.5
& 0.98B
& COCO & -- & 77.6 & 80.7 \\

SDPose (Ours)
& Stable Diffusion-v1.5
& 0.98B
& COCO & 40 & 81.2 & \textbf{85.3} \\

SDPose (Ours)
& Stable Diffusion-v2
& 0.98B
& COCO & 40 & \textbf{81.3} & 85.2 \\
\noalign{\hrule height 1.2pt}
\end{tabular}
\end{adjustbox}

\end{table}

\begin{figure}[!t]
    \centering
    \includegraphics[width=0.65\linewidth,height=0.30\textheight,keepaspectratio]{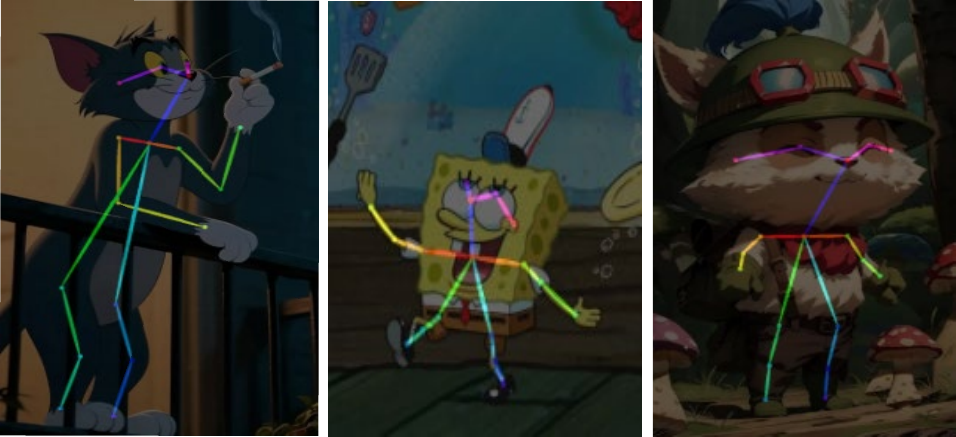}
    \caption{\textbf{Failure Cases of SDPose.} SDPose may produce inaccurate keypoints on anthropomorphic or non-human characters, especially around ears, facial landmarks, and body parts with substantially different proportions from humans.}
    \label{fig:failure_cases}
\end{figure}

\subsection{Inference Cost}
\label{sec:inference_cost}

We compare the inference cost of SDPose with representative pose estimation methods in Table~\ref{tab:tflops}. At the same $1024{\times}768$ resolution, SDPose requires fewer parameters and FLOPs than Sapiens-1B/2B, while maintaining competitive COCO performance and stronger HumanArt performance.

\begin{table}[!t]
\caption{\textbf{Inference cost and accuracy comparison.}
TFLOPs exclude person detection. ViTPose-H is trained on COCO+AIC+MPII; the remaining models are trained or fine-tuned on COCO only.}
\label{tab:tflops}
\centering
\renewcommand{\arraystretch}{1.1}
\setlength{\tabcolsep}{1.5pt}

\begin{adjustbox}{max width=\linewidth,center}

\begin{tabular}{lccccccc}
\noalign{\hrule height 1.2pt}
\textbf{Method} &
\textbf{Kpts} &
\textbf{Res.} &
\textbf{Params.} &
\textbf{TFLOPs} &
\textbf{FPS} &
\textbf{COCO} &
\textbf{HumanArt} \\
\hline
Sapiens-Pose-1B & Body &
$1024{\times}768$ & 1.17B & 4.88 & 13.59 &
82.1/85.9 & 64.3/67.4 \\

Sapiens-Pose-2B & Body &
$1024{\times}768$ & 2.20B & 8.52 & 8.04 &
\textbf{82.2/86.0} & 69.6/72.2 \\

SDPose (Ours) & Body &
$1024{\times}768$ & 0.98B & 3.62 & 9.20 &
81.2/85.3 & \textbf{71.8/74.6} \\

ViTPose-H & Body &
$256{\times}192$ & 0.64B & 0.126 & 200.18 &
79.1/84.1 & 66.5/70.1 \\
\hline
Sapiens-Pose-1B & WholeBody &
$1024{\times}768$ & 1.17B & 4.88 & 13.51 &
72.7/79.2 & -- \\

Sapiens-Pose-2B & WholeBody &
$1024{\times}768$ & 2.20B & 8.52 & 8.00 &
74.4/81.0 & -- \\

SDPose (Ours) & WholeBody &
$1024{\times}768$ & 0.98B & 3.62 & 8.94 &
72.8/79.4 & -- \\

DWPose & WholeBody &
$384{\times}288$ & 33.61M & 0.01 & 135.23 &
66.5/74.3 & -- \\
\noalign{\hrule height 1.2pt}
\end{tabular}
\end{adjustbox}
\end{table}

At the matched $1024{\times}768$ resolution, SDPose requires 3.62 TFLOPs, compared with 4.88 and 8.52 TFLOPs for Sapiens-1B and Sapiens-2B, respectively. ViTPose-H and DWPose achieve substantially higher throughput at lower input resolutions, reflecting a different accuracy--efficiency trade-off.

\section{Additional Strong Baseline Comparisons}
\label{sec:supp_strong_baselines}

To further disentangle the effects of generic large-scale visual pre-training and input resolution, we introduce two strong controlled baselines, DINOv2-L and ViTPose-H. Table~\ref{tab:strong_baselines} summarizes their performance on COCO, HumanArt, and COCO-OOD Monet.

\begin{table}[t]
\caption{\textbf{Controlled comparison with strong visual representation baselines.}
All models are trained on COCO and evaluated at the same $1024\times768$ input resolution.}
\label{tab:strong_baselines}
\centering
\renewcommand{\arraystretch}{1.15}
\setlength{\tabcolsep}{4pt}
\footnotesize

\begin{adjustbox}{max width=\linewidth,center}

\begin{tabular}{lcc cc cc cc}
\toprule
\textbf{Model}
& \textbf{Backbone}
& \textbf{Input Res.}
& \multicolumn{2}{c}{\textbf{COCO}}
& \multicolumn{2}{c}{\textbf{HumanArt}}
& \multicolumn{2}{c}{\makecell[c]{\textbf{COCO-OOD}\\\textbf{Monet}}} \\
\cmidrule(lr){4-5}
\cmidrule(lr){6-7}
\cmidrule(lr){8-9}

& &
& \textbf{AP} & \textbf{AR}
& \textbf{AP} & \textbf{AR}
& \textbf{AP} & \textbf{AR} \\
\midrule

ViTPose-H~\cite{xu2022vitpose}
& MAE ViT-H
& $1024\times768$
& 80.4 & 84.5
& 67.4 & 70.4
& 61.7 & 66.6 \\

DINOv2-L~\cite{oquab2023dinov2}
& DINOv2 ViT-L
& $1024\times768$
& 79.8 & 83.9
& 66.4 & 69.8
& 59.3 & 64.1 \\

\textbf{SDPose (Ours)}
& Stable Diffusion-v2
& $1024\times768$
& \textbf{81.2} & \textbf{85.3}
& \textbf{71.8} & \textbf{74.6}
& \textbf{64.0} & \textbf{68.7} \\

\bottomrule
\end{tabular}
\end{adjustbox}
\end{table}

\paragraph{DINOv2-L.}
For a controlled comparison of visual pre-training paradigms, we evaluate DINOv2-L with the same heatmap head against our single-feature $F_1$ + heatmap head baseline under the same $1024\times768$ input resolution. DINOv2-L achieves \textbf{79.8/83.9}, \textbf{66.4/69.8}, and \textbf{59.3/64.1 AP/AR} on COCO, HumanArt, and COCO-OOD Monet, respectively, while our $F_1$ baseline obtains \textbf{81.3/85.2}, \textbf{71.2/73.9}, and \textbf{63.5/68.2}. The AP gap increases from \textbf{+1.5} on in-domain COCO to \textbf{+4.8/+4.3} on HumanArt/Monet, supporting stronger cross-domain transferability enabled by SDPose's exploitation of diffusion representations beyond generic large-scale self-supervised visual pre-training.

\paragraph{ViTPose-H at Matched Resolution.}
We further train ViTPose-H at $1024\times768$, matching the high-resolution setting of SDPose. Increasing its resolution from $256\times192$ to $1024\times768$ improves COCO AP from \textbf{79.1} to \textbf{80.4} and HumanArt AP from \textbf{66.5} to \textbf{67.4}. Under this matched-resolution setting, ViTPose-H achieves \textbf{80.4/84.5}, \textbf{67.4/70.4}, and \textbf{61.7/66.6 AP/AR} on COCO, HumanArt, and COCO-OOD Monet, respectively, while SDPose achieves \textbf{81.2/85.3}, \textbf{71.8/74.6}, and \textbf{64.0/68.7}. SDPose therefore maintains gains of \textbf{+0.8}, \textbf{+4.4}, and \textbf{+2.3 AP} across the three datasets. These results show that increased input resolution improves ViTPose-H's in-domain accuracy but does not account for SDPose's stronger cross-domain performance on either HumanArt or COCO-OOD Monet.

\section{Failure Case Study}

As shown in Fig.~\ref{fig:failure_cases}, SDPose mainly fails on anthropomorphic and non-human characters whose anatomy deviates from humans. Typical errors include inaccurate ear localization and incorrect body keypoints on characters with unusual proportions or human-like poses. We attribute these failures mainly to the human-only downstream fine-tuning data, which limits the transfer of human keypoint semantics to non-human structures.

\section{COCO-OOD Human Baseline}

Three annotators independently relabel 100 COCO-OOD images. The high Human OKS
(\textbf{0.8937}) and inter-annotator agreement ($P_o=0.964$, $\kappa=0.741$)
confirm that style transfer largely preserves pose geometry and the validity of
the inherited COCO annotations.

\begin{table}[t]
\centering
\small
\caption{Human baseline on COCO-OOD, averaged over all 17 keypoints and over three independent annotators. $P_o$ and $\kappa$ are the raw agreement and Fleiss' $\kappa$ of the binary \emph{placed / not placed} decision across the three annotators.}
\label{tab:human_baseline}
\begin{tabular}{l c r rr}
\toprule
Split & $n$ & OKS & $P_o$ & $\kappa$ \\
\midrule
COCO-Monet   & 50  & 0.8975 & 0.960 & 0.733 \\
COCO-Ukiyo-e & 50  & 0.8899 & 0.967 & 0.749 \\
\midrule
All          & 100 & 0.8937 & 0.964 & 0.741 \\
\bottomrule
\end{tabular}
\end{table}
}
\end{document}